\documentclass[a4paper, 10pt, conference]{ieeeconf}      

\IEEEoverridecommandlockouts                              
\usepackage{amsmath}

\usepackage{amsthm}

\usepackage{graphics} 
\usepackage{epsfig} 
\usepackage{amssymb}  
\usepackage{multicol}
\usepackage{subcaption}
\usepackage{graphicx}
\usepackage[utf8]{inputenc} 
\usepackage[T1]{fontenc}    
\usepackage{hyperref}       
\usepackage{url}            
\usepackage{booktabs}       
\usepackage{amsfonts}       
\usepackage{nicefrac}       
\usepackage{microtype}      
\usepackage{xcolor}         
\usepackage{dsfont}

\usepackage{mathtools}
\usepackage[linesnumbered,ruled]{algorithm2e}

\newtheorem{theorem}{Theorem}[section]

\newtheorem{lemma}[theorem]{Lemma}

\newtheorem{remark}[theorem]{Remark}

\makeatletter
\def\BState{\State\hskip-\ALG@thistlm}
\makeatother

\newcommand{\ones}{\mathds{1}}
\newcommand{\tr}{\top}

\newcommand{\Rx}{R_{\textnormal{max}}}
\newcommand{\cA}{\mathcal{A}}

\newcommand{\bE}{\mathbb{E}}

\newcommand{\cF}{\mathcal{F}}

\newcommand{\cM}{\mathcal{M}}

\newcommand{\bR}{\mathbb{R}}
\newcommand{\cP}{\mathcal{P}}
\newcommand{\cR}{\mathcal{R}}
\newcommand{\cS}{\mathcal{S}}
\newcommand{\cT}{\mathcal{T}}
\newcommand{\cU}{\mathcal{U}}

\newcommand{\KA}{C_A}
\newcommand{\KAi}{C_{A, \mathrm{inv}}}
\newcommand{\Kb}{C_b}
\newcommand{\Kd}{C_d}

\newcommand{\KM}{C_{M}}
\newcommand{\KMr}{C_{M, r}}
\newcommand{\KMt}{C_{M, \theta}}

\newcommand{\Ht}{H_\theta}
\newcommand{\Hr}{H_r}

\newcommand{\ep}{\epsilon_p}
\newcommand{\er}{\epsilon_r}

\newcommand{\dm}{d^\mu}
\newcommand{\Pm}{\cP^\mu}
\newcommand{\Rm}{\cR^\mu}
\renewcommand{\rm}{r^\mu}

\newcommand{\thS}{\theta^*}
\newcommand{\rS}{r^*}

\newcommand{\diag}{\textnormal{diag}}

\title{\LARGE \bf Federated TD Learning in Heterogeneous Environments with Average Rewards: A Two-timescale Approach with Polyak-Ruppert Averaging
}

\author{Ankur Naskar$^a$, Gugan Thoppe$^a$, Abbasali Koochakzadeh$^b$, and Vijay Gupta$^b$ 
\thanks{ }
\thanks{$^a$A.~Naskar ({\tt\small ankurnaskar@iisc.ac.in}) and G.~Thoppe ({\tt\small gthoppe@iisc.ac.in}) are with the Department of Computer Science and Automation, Indian Institute of Science, Bengaluru, India }%
\thanks{$^b$A.~Koochakzaadeh ({\tt\small akoochak@purdue.edu}) and V.~Gupta ({\tt\small gupta869@purdue.edu}) are with the Department of Electrical and Computer Engineering, Purdue University, IN, USA}%
}

\begin{document}

\maketitle
\thispagestyle{empty}
\pagestyle{empty}

\begin{abstract}
Federated Reinforcement Learning (FRL) provides a promising way to speedup training in reinforcement learning when one has access to multiple edge devices that can operate in parallel. Recently, it has been shown that even when these edge devices have access to different dynamic models---such as cars or robots with differing system configurations---an optimal convergence rate that has a linear speedup proportional to the number of devices is achievable. However, this result requires that the algorithm's stepsize be chosen in a manner dependent on the unknown model parameters. Also, it applies only to a discounted setting, which has been argued to fit episodic tasks better than continuing control tasks. In this paper, we obtain finite-time bounds for heterogeneous FRL with average rewards. We show that the optimal convergence rate and the linear speedup is possible even with a universal stepsize choice, independent of the underlying dynamics. To achieve our result, we modify the existing one-timescale FRL method to a novel two-timescale variant that additionally incorporates iterate averaging. We remark that this idea also extends to other settings already studied in the literature, such as the homogeneous FRL or the heterogeneous case with a discounted cost function.








\end{abstract}

\section{Introduction}
\label{sec:intro}

With the proliferation of edge devices such as scanners, smartphones, medical devices, scientific instruments, and autonomous vehicles, there has been a massive growth in the amount of data that could conceivably be used for training Machine Learning (ML) models. However, transmitting this data from the edge devices to a single centralized location can easily overwhelm the network bandwidth. It can also raise privacy concerns, as the edge devices may not wish to reveal sensitive attributes embedded in their data. The paradigm of Federated Learning (FL) has evolved to address these concerns by allowing model training to be done primarily by the edge devices themselves~\cite{mcmahan2017communication, bonawitz2019towards}. 

Fundamentally, FL consists of three steps that are repeated in a cyclic manner. First, the edge devices locally train an ML model. Next, the server gathers these models, possibly by sampling the devices, and then aggregates them. Finally, the server transmits this global model to the edge devices. The core challenge of FL is that the edge devices and their contributions may be heterogeneous. The most obvious form of this heterogeneity is that of data heterogeneity, meaning that the edge devices operate in distinct environments and collect data from distributions that are not identical to each other. Much of the literature in FL accordingly provides ways to combat such heterogeneity.

A branch of ML that has seen a surge of interest recently is Reinforcement Learning (RL) \cite{sutton2018reinforcement, levine2020offline, recht2019tour, cheng2021general, kyriakis2019specification}, wherein an agent is required to learn an optimal strategy (or a policy) for a control system. Formally, a Markov Decision Process (MDP)  models the system dynamics such that the (probability distribution of the) future state of the system depends only on its current state and an action performed by an agent. After each action, the agent receives a scalar feedback, called the reward, that quantifies the immediate gains under that action. The overall goal in RL then is to find a policy for the agent that optimizes a suitable cumulative sum of the instantaneous gains. We remark that RL has been successfully used across a diverse set of applications including robot locomotion \cite{DBLP:journals/corr/SchulmanWDRK17}, the Atari \cite{NIPS2014_8bb88f80} and Go games \cite{Silver2016}, autonomous vehicles, resource management in computer clusters \cite{mao2016resource}, traffic light control
\cite{arel2010reinforcement},  web system configuration \cite{bu2009reinforcement}, and personalized recommendations
\cite{zheng2018drn}.

There are two main ways in which the above cumulative sum (of rewards) has been defined in the literature. One way is to look at a discounted sum of the instantaneous rewards (with some discount factor $\gamma \in [0,1)$) over some time horizon. However, it has been argued that such discounting is suitable for episodic tasks but leads to high near-term performance rather than to high long-term performance in continuing or infinite-horizon control tasks~\cite{sutton1999policy, naik2019discounted, zhang2021finite}. The alternative is the average-reward setting, the focus of this work. 

Some recent works have merged the FL and RL paradigms and studied Federated Reinforcement Learning (FRL)~\cite{qi2021federated,nadiger2019federated,zhuo2019federated,wang2020optimizing}. One advantage of FRL is that a `divide and conquer' approach---to generate the usually enormous amount of data required to explore the different aspects of the environment---becomes conceivable. Moreover, the different edge devices can now perform  computations in parallel, leading to a possible speedup in terms of the number of participating devices. Initial works show that this intuition is true at least in the case when each node has access to the same model for the system process.

However, as stated above, heterogeneity is a key concern in FL. In FRL, this may manifest as different edge devices working with (slightly) different dynamic models. For instance, to design a controller for an autonomous car, we can try to utilize data from several cars; however, every car operates in a different environment with different configurations. Similarly, even if multiple robots are operated under similar conditions, it is not possible to have all the parameters in their physical models to be identical. It is important to note that, due to the global aggregation step in FL, the designed controller may end up not being optimal for the dynamic model at any single edge device. Hence, one typically works with an alternative goal in heterogenous FRL, which is to find a `universal' controller that performs well across all the edge models. An intriguing question is whether the speedup from the homogeneous model case can be achieved in the heterogeneous case as well for finding this universal controller. 

Some recent works~\cite{jin2022federated, wang2023federated} have studied the above question. Specifically,~\cite{wang2023federated} has considered policy evaluation with linear function approximation using Temporal-Difference (TD) learning and proved that the optimal convergence rate and a linear speedup persists even with heterogeneity. However, their setup is of discounted rewards, which as stated above, is less suitable for continuing control tasks than  average rewards. Moreover, for  the optimal rate, the algorithm
in~\cite{wang2023federated} is required to choose stepsizes in a manner that depend on the unknown dynamic models at the edge devices, which is practically infeasible.

In this work, we consider heterogeneous FRL with an average-reward criterion and show that the optimal rate and the linear speedup are still possible. More significantly, we show that this speedup can be achieved by choosing stepsizes that do not depend on the unknown dynamics at the edge devices. A formal summary of our key contributions is as follows. We propose a novel \textit{two-timescale} variant of TD learning, complemented with the \textit{Polyak-Ruppert averaging}, for policy evaluation in heterogeneous MDP settings with the \textit{average-reward} criterion and linear function approximation.  Our main result shows that our algorithm achieves the \textit{optimal rate} with a \textit{linear speedup} in iteration complexity in terms of the number of agents. Importantly, we show that this  optimality occurs for a \textit{universal} stepsize choice that is independent of the underlying dynamics. 

The rest of the paper is organized as follows. We begin with formulating the problem in Section~\ref{sec:setup}. Our proposed algorithm and our main result (Theorem~\ref{thm : finite-time convergence}) are given in Section~\ref{sec:algorithm}. We prove our main result in Section~\ref{sec:proof} and conclude with some open directions in Section~\ref{sec:future}.

\section{Setup and Problem Formulation}
\label{sec:setup}

Our framework comprises $N$ agents (also referred to as clients or nodes), with the $i$-th agent having access to an MDP $\cM_{i}:= (\cS, \cA, \cR_{i}, \cP_{i}).$  Here, $\cS$ and $\cA$ are the state and action spaces, respectively, and are assumed to be finite and common among all the MDPs. Further,  $\cR_{i}:\cS\times \cA\to \bR$ and $\cP_{i}:\cS\times \cA\to \Delta (\cS)$ are the local reward and probability transition functions at agent $i \in [N]$ and can potentially vary from one agent to the other. The notation $\Delta(\cU)$ stands for the set of distributions on a finite set $\cU,$ and $[N] := \{1, \ldots, N\}.$ Our goal is to design an FRL algorithm that leverages the above setup to estimate the value function of a stationary policy $\mu: \cS \to \Delta(\cA)$ under the average-reward criterion. In particular, our algorithm's output should approximate $\mu$'s value function in the column space of a given feature matrix $\Phi \in \bR^{|\cS| \times d}$ for some $1 \leq d \leq |\cS|.$

We now formally define $\mu$'s value function with respect to the MDP $\cM_i$ under the average-reward criterion. It is measured using two notions: the average reward and the differential value function. The average reward $\rm_i \in \bR^{|\cS|}$ is given by
\begin{equation}\label{e: avg.reward}
    \rm_i(s) := \liminf\limits_{T\to\infty}\frac{1}{T} \bE \bigg[\sum_{t = 0}^{T - 1} \cR_i(s_{t}, a_t) | s_{0} = s \bigg], \quad s \in \cS,
\end{equation}
where the expectation is with respect to the distribution of the state-action trajectory $s_0, a_0,  \ldots, s_{T - 1}, a_{T - 1},$ in which $s_0 = s,$ and $a_t \sim \mu(\cdot|s_t)$ and $s_{t + 1} \sim \cP_i(\cdot|s_t, a_t)$ for $t \geq 0.$ In contrast, the differential value function $V^\mu_i$ is the fixed point of the differential Bellman operator $T^\mu_i: \bR^{|\cS|} \to \bR^{|\cS|}$ which is given by
\begin{equation}
    T^\mu_i V = \Rm_i - \rm_i + \Pm_i V.
\end{equation}
Above, $\Rm_i(s) = \sum_{a \in \cA} \mu(a|s) \cR_i(s, a),$ and $\Pm_i(s, s') \equiv \Pm_i(s'|s) = \sum_{a \in \cA} \mu(a|s) \cP_i(s'|s, a).$ Throughout this work, we presume that the data (i.e, the states and actions) observed at different agents are independent. Additionally, we make the following standard assumption \cite{zhang2021finite,wang2023federated}: 
{
\renewcommand{\theenumi}{$\mathbf{\cA_\arabic{enumi}}$}
\begin{enumerate}   

     \item\label{a: ergodic} \textbf{Ergodicity}: For any $i \in [N],$ the Markov chain $(\cS, \cP^\mu_i)$ induced by the policy $\mu$ is irreducible and aperiodic.
\end{enumerate}
}
\noindent This latter assumption guarantees that the Markov chain $(\cS, \cP^\mu_i)$ has a unique and positive stationary distribution $\dm_i;$ further, this Markov chain is ergodic, and, for each $s \in \cS,$ 
\begin{equation}\label{e:avg.reward}
    \rm_i(s) = (\dm_i)^\tr \Rm_i =: \rS_i.
\end{equation}
Note that $\rS_i$ and, hence, $\rm_i(s)$ is independent of $s.$

In the above notations, our FRL algorithm's goals can be restated as follows. It should output a vector $\theta \in \bR^d$ and a scalar $r \in \bR$ such that $\Phi \theta$ (resp. $r$) is simultaneously close to $V^\mu_i$ (resp. $\rS_i$) for each $i.$ Moreover, our algorithm's convergence rate should be optimal and its iteration complexity---i.e., the number of iterations needed to find such a $\theta$ and $r$---should decrease linearly with the number $N$ of agents. Finally, the stepsize choice to achieve this optimal rate should not depend on unknown problem dependent constants. 

\section{Proposed Algorithm and Main Result}
\label{sec:algorithm}


\let\oldnl\nl
\newcommand{\nonl}{\renewcommand{\nl}{\let\nl\oldnl}}

\begin{algorithm}[t]
\caption{Our proposed AvgFedTD(0) algorithm\label{alg: FedTD}}
\SetKwInOut{Input}{Input}
\SetKwFor{For}{for}{}{end}
\SetKwBlock{Central}{Central server}{}
\SetKwBlock{Local}{Each agent $i\in [N]$ in parallel}{}
\Input{Policy $\mu,$ step-size sequence $(\beta_{t}),$ feature vectors $\{\phi(s): s\in \cS\},$ initial average reward estimate $r_{0} \in \bR,$ initial global model parameter $\theta_{0}\in \bR^{d}.$} 

\textbf{Initialize:} $\bar{\theta}_{0} = \theta_{0},$ and $r_0^i=r_0,\forall i\in [N].$

\nonl \For{each iteration $t= 0,1, \ldots, T-1:$}
   { 
        \nonl\Local
        {
            Observe $(s_{t}^{i},a_{t}^{i},\hat{s}_{t}^{i}),$ 
            where $s^{i}_{t}\sim d_{i}^{\mu},$ 
            $a_{t}^{i}\sim\mu(\cdot|s_{t}^{i}),$ and $\hat{s}_{t}^{i} \sim P_{i}(\cdot|s_{t}^{i}, a_{t}^{i}).$

            \vspace{1ex}
         
            Compute local TD error 
            $\delta_{t+1}^{i} = [\cR_{i}(s_{t}^{i},a_{t}^{i}) - r_{t}] \phi(s_{t}^{i})$
            
            \nonl\hspace{1.5cm} $+\ \phi(s_{t}^{i})[\phi^{\tr}(\hat{s}_{t}^{i}) - \phi^{\tr} (s_{t}^{i})]\theta_{t}.$ 

            \vspace{1ex}
            
            Update local average reward estimate  $r_{t+1}^{i} =  r_{t}^{i} + \frac{1}{t+1}[\cR_{i}(s_{t}^{i},a_{t}^{i}) - r_{t}^{i}].$ \nllabel{alg:local.reward.update}

            \vspace{1ex}

            Send $(\delta_{t+1}^{i}, r^{i}_{t+1})$ to central server.

        }

        \nonl\Central
        {
            \nllabel{alg:global.parameter.update} Update global model parameter
            $\theta_{t+1} = \theta_{t} + \frac{\beta_{t}}{N}\sum_{i\in[N]}\delta_{t+1}^{i}.$
            
            \vspace{1ex}
            
            Update Polyak-Ruppert average  
            $\bar{\theta}_{t+1}  = \bar{\theta}_{t} +   \frac{1}{t+1}[\theta_{t} - \bar{\theta}_{t}].$ \nllabel{alg:PR}

            \vspace{1ex}
            
            Update average reward estimate
            $r_{t+1} = \frac{1}{N}\sum_{i\in[N]}r_{t+1}^{i}.$

            \vspace{1ex}
            
            Send $(\theta_{t+1},r_{t+1})$ to each agent $i\in[N].$
        }
   }
\end{algorithm}

Our novel algorithm for policy evaluation in the heterogeneous FL setup under the average reward criterion is given in Algorithm~\ref{alg: FedTD}. There are two phases in each iteration of this algorithm and their details are as follows. In the first phase (Steps 2--5), all agents work in parallel to compute their local TD errors and their local estimates of the average reward, i.e., the $\delta^i_t$'s and the $r^i_t$'s, which are then shared with a central server. In the second phase (Steps 6--9), the server aggregates these quantities to obtain global estimates for the differential value function and the average reward, which are then broadcast to all the agents. To compute the local TD error $\delta^i_t,$ we assume that agent $i$ has access to the current state $s^i_t \sim d^\mu_i$ of its local MDP $\cM_i,$ and can take action $a^i_t \sim \mu(s^i_t),$ after which it gets to see the subsequent local state $\hat{s}^i_{t} \sim \cP^\mu_i(\cdot|s^i_t, a^i_t),$ and the local instantaneous reward $\cR_i(s^i_t, a^i_t).$ Specifically, we presume that $s^i_t, a^i_t, \hat{s}^i_{t}$ can be obtained with independent randomness\footnote{In this work, we only focus on the generative model, i.e., the case where $(s^i_t) \overset{IID}{\sim} d^\mu_i.$ The extension of our algorithm and results to the Markovian sampling case can be found in our follow up work in \cite{naskar2025parameterfree}.}. Note that the information about the local states, actions, and rewards is not shared directly with the server, following FL principles. 

We now discuss how our algorithm differs from \cite[Algorithm~1]{wang2023federated} (which concerns the \textit{discounted} case) and also the latter's limitation. For ease of comparison, we first describe a naive extension of the latter algorithm to the average-reward setting. This extension is obtained by incorporating in \cite[Algorithm~1]{wang2023federated} the ideas from \cite[Algorithm~1]{zhang2021finite}, which is designed for policy evaluation in the \textit{single agent} and \textit{average-reward} setting. The resultant algorithm resembles Algorithm~\ref{alg: FedTD}, but with a few changes:
\begin{enumerate}
    \item In Step~\ref{alg:local.reward.update}, the $1/(t + 1)$ stepsize is replaced by $c_\alpha \beta_t;$ 

    \item Step~\ref{alg:PR} is omitted.
\end{enumerate}
Regarding limitation, it can be seen from \cite[Theorem~1]{zhang2021finite}  that the optimal convergence rates for $\theta_t$ and $r_t$ in this naive extension---which, in turn, have a linear speedup in terms of iteration complexity---requires stepsizes to depend on unknown problem parameters. In particular, in \cite{zhang2021finite}, the stepsize $\beta_t$ is chosen of the form $c_1/(c_2 + t)$ and the constants $c_1, c_2,$ and $c_\alpha$ are such that $2 < c_1 \Delta < 2 c_2$ and $c_\alpha > \Delta + \sqrt{\Delta^{-2} - 1},$ where $\Delta:=\min_{\theta\in\bR^{d}: \|\theta\| = 1} \theta^{\tr} A\theta$ and $A = \frac{1}{N} \sum_{i \in [N]} A_i$ with $A_{i}:= \Phi^{\tr}D_{i}^{\mu}(I - \cP_{i}^{\mu})\Phi$ and $D_{i}^{\mu}:= 
\diag(d_{i}^{\mu})$ for all $i \in [N].$ Clearly, this stepsize choice is impractical, as it necessitates the knowledge of $\Delta$ which depends on $\cP^\mu_i,$ an a priori unknown matrix. 

Our approach differs from the aforementioned naive algorithm in the following two crucial ways. The first is that we have the additional Step~\ref{alg:PR}, wherein we perform a \textit{Polyak-Ruppert} \cite{polyak1992acceleration} running averaging of the global $\theta$-estimates, i.e., we compute $\bar{\theta}_t = \frac{1}{t}\sum_{k = 0}^{t - 1} \theta_k$ at each time instance $t.$ The second is in changing the updating approach of $\theta_t$ and $r^i_{t}$ from one-timescale to \textit{two-timescales}. Put differently, in the naive extension,  both $\theta_t$ and $r^i_t$ are updated using stepsizes that differ only by a constant factor. In contrast, in our algorithm, we update these quantities using the stepsizes $\beta_t$ and $1/(t + 1),$ respectively, and we assume that their ratio goes to $\infty$ as $t \to \infty$. As an example, this condition holds for $\beta_t = 1/(t + 1)^\beta$ when $\beta$ is some constant in $(0, 1).$

Because of the above two modifications, we now show that our algorithm does not have the impracticality issue inherent in the above naive algorithm. That is, it achieves the optimal convergence rate and a linear speedup in the associated sample complexity without requiring the stepsize $\beta_t$ to depend on unknown problem-dependent parameters.  

We need the following assumptions to formally state our main result. Let $\|\cdot\|$ be the Euclidean norm.
{
\renewcommand{\theenumi}{$\mathbf{\cA_\arabic{enumi}}$}
\begin{enumerate}  

\setcounter{enumi}{1}

     
     \item\label{a: dynamics.hetero} \textbf{Heterogeneity bound}:  $\exists\ \ep \in [0, 1)$ and $\er >0$ such that 
     \[
        |\cR_{i}(s, a) - \cR_{j}(s, a)| \leq \er
    \]
    and 
     \[ 
        \lvert \cP_{i}(s'|s, a) - \cP_{j}(s'|s, a) \rvert \leq \ep\ \cP_i(s'|s, a)
    \]
    $\forall i,j \in [N],$ $ s\neq s'\in \cS,$ and $a \in \cA.$ 

    \item\label{a: reward.bound} \textbf{Bounded rewards}:  $\exists \Rx > 0$ such that $\lvert \cR_{i}(s,a)\rvert \leq \Rx$ $\forall i \in [N], \forall s\in \cS,$ and $\forall a \in \cA.$

    \item\label{a: feature.matrix} \textbf{Conditions on the feature matrix}: The matrix $\Phi$ has full-column rank with $\|\Phi\| \leq 1.$ Additionally, the column space of $\Phi$ does not contain the vector of all ones, i.e., $\ones\notin \{\Phi\theta :\theta \in \bR^{d}\}.$
    
 \end{enumerate}   
}

\noindent Assumption \ref{a: dynamics.hetero} puts an upper bound on the heterogeneity among the local environments. Separately, Assumption\ref{a: feature.matrix}, which is also used in \cite{zhang2021finite}, along with \ref{a: ergodic} is required for
the positive definiteness of each $A_i$ \cite[Lemma~2]{zhang2021finite}. The positive definiteness of $A$ follows from that of the $A_i$'s.

We also need a few notations. For all $i \in [N],$ let $b_{i} := \Phi^{\tr}D_{i}^{\mu}\cR_{i}^{\mu},$ $v_{i}  := \Phi^{\tr}D_{i}^{\mu}\ones,$ and $\theta_{i}^{*} := A_{i}^{-1}(b_{i} - v_{i}r^{*}_{i}).$ Separately, let $b := \frac{1}{N} \sum_{i\in [N]} b_{i},$ $v := \frac{1}{N}\sum_{i\in [N]} v_{i},$ and $r^{*}:= \frac{1}{N}\sum_{i\in[N]} r_{i}^{*}.$ Finally, let $\theta^{*} := A^{-1}(b - vr^{*}).$ 

We are now ready to state our main result on the finite-time convergence of our AvgFedTD(0) algorithm.

\begin{theorem}
\label{thm : finite-time convergence}
Suppose Assumptions~\ref{a: ergodic}--\ref{a: feature.matrix}
hold. Let $(\bar{\theta}_t,r_t)$ be the iterates
generated by Algorithm~\ref{alg: FedTD}, with
$\beta_t=(t+1)^{-\beta}$ for some $\beta\in(0,1)$.
Then, for every $i\in[N]$ and $T\geq1$, we have
\begin{align*}
\bE|r_T-r_i^*|^2
&\leq \frac{\Rx^2}{NT}
+\Hr(\ep,\er),
\intertext{and}
\bE\|\bar{\theta}_T-\theta_i^*\|^2
&\leq
\frac{2C_{\bar{\Delta}}^{(0)}}{T^2}
+\frac{2C_{\bar{\Delta}}^{(1)}}{NT}
+2\Ht(\ep,\er),
\end{align*}
where $C_{\bar{\Delta}}^{(0)}$,
$C_{\bar{\Delta}}^{(1)}$,
$\Hr(\ep,\er)$, and $\Ht(\ep,\er)$
are as defined in Table~\ref{tab: constants.FedTD}.
\end{theorem}

\begin{remark}
Theorem~\ref{thm : finite-time convergence} shows
that the estimation error relative to the federated
solution $(\theta^*,r^*)$ decays at the rates
$O(T^{-2}+(NT)^{-1})$ and $O((NT)^{-1})$,
respectively. In particular, the stochastic
estimation error exhibits a linear reduction
with the number of agents, provided the
problem-dependent constants remain uniformly
bounded as $N$ varies.

In heterogeneous environments, however, the
federated solution need not coincide with the
individual agents' solutions. Indeed, our
convergence results imply that, for every
$i\in[N]$,
\[
\begin{aligned}
\lim_{T\to\infty}
\bE|r_T-r_i^*|^2
&=|r^*-r_i^*|^2
\leq \Hr(\ep,\er),\\
\lim_{T\to\infty}
\bE\|\bar{\theta}_T-\theta_i^*\|^2
&=\|\theta^*-\theta_i^*\|^2
\leq \Ht(\ep,\er).
\end{aligned}
\]
Thus, collaboration reduces the stochastic
estimation error but may introduce a persistent
error relative to an individual agent's solution.
This tradeoff is consistent with the observations
in~\cite{wang2023federated}.
\end{remark}

\begin{remark}
\label{rem:benefits}
A key advantage of our algorithm is that it
achieves these convergence rates using the
stepsize $\beta_t=(t+1)^{-\beta}$ for any
$\beta\in(0,1)$, without requiring knowledge
of the underlying transition matrices or
problem-dependent spectral quantities.

This contrasts with the finite-time analyses
in~\cite{zhang2021finite,wang2023federated},
which establish their convergence guarantees
using stepsize choices that depend on
problem-specific parameters. In our approach,
such parameters affect the constants in the
convergence bounds but not the implementation
of the algorithm.
\end{remark}

\begin{table}[t!]
\centering
\caption{Constants used in the finite-time analysis.}
\label{tab: constants.FedTD}
\footnotesize
\renewcommand{\arraystretch}{1.12}
\begin{tabular}{|l|l|}
\hline
\textbf{Constant} & \textbf{Definition} \\ \hline

\multicolumn{2}{|l|}{\textit{Heterogeneity and stability}} \\ \hline

$\Kd(\ep)$ &
$\displaystyle
\left(\frac{1+\ep}{1-\ep}\right)^{|\cS|}-1$ \\

$\KAi$ &
$\displaystyle
\max\left\{
\max_{i\in[N]}\|A_i^{-1}\|,
\|A^{-1}\|
\right\}$ \\

$\KA(\ep)$ &
$\displaystyle
2\min\{2,\Kd(\ep)\}+2\ep$ \\

$\Kb(\ep,\er)$ &
$\displaystyle
2\er+3\Rx\min\{2,\Kd(\ep)\}$ \\

$\Hr(\ep,\er)$ &
$\displaystyle
\big[\er+\Rx\min\{2,\Kd(\ep)\}\big]^2$ \\

$\Ht(\ep,\er)$ &
$\displaystyle
\big[
\KAi\Kb(\ep,\er)
+\KAi^2\KA(\ep)\Rx
\big]^2$ \\ \hline

\multicolumn{2}{|l|}{\textit{Noise and one-step convergence}} \\ \hline

$\KM$ &
$\displaystyle 96(\Rx^2+\|\thS\|^2)$ \\

$\KMr$ &
$\displaystyle 12$ \\

$\KMt$ &
$\displaystyle 48$ \\

$\lambda$ &
$\displaystyle
\text{any fixed number in }
(0,\lambda_{\min}(A+A^\tr))$ \\

$t_\Delta$ &
$\displaystyle
\max\left\{
1,
\left\lceil
\left(\frac{4(4+\KMt)}{\lambda}\right)^{1/\beta}
\right\rceil
\right\}$ \\

$C_\Delta^{(2)}$ &
$\displaystyle
\KM+
\left(1+\KMr+\frac{36}{\lambda}\right)\Rx^2$
\\ \hline

\multicolumn{2}{|l|}{\textit{Mean-square convergence}} \\ \hline

$a$ &
$\displaystyle \frac{\lambda}{4}$ \\

$t_*$ &
$\displaystyle
\max\left\{
t_\Delta,
\left\lceil
\left(\frac{8\beta}{\lambda}\right)^{1/(1-\beta)}
\right\rceil
\right\}$ \\

$D$ &
$\displaystyle \frac{8C_\Delta^{(2)}}{\lambda}$ \\

$B$ &
$\displaystyle 5+\KMt$ \\

$U_1$ &
$\displaystyle
(18+\KMt)\bE\|\Delta_0\|^2
+(2+\KMr)\bE\rho_0^2+\KM$ \\

$U_*$ &
$\displaystyle
B^{t_*-1}
\left(U_1+\frac{C_\Delta^{(2)}}{B-1}\right)$ \\

$c$ &
$\displaystyle \frac{\lambda}{4(1-\beta)}$ \\

$C_\Delta^{(0)}$ &
$\displaystyle
U_*\exp\big(c(t_*+1)^{1-\beta}\big)$ \\

$C_\Delta^{(1)}$ &
$\displaystyle D$ \\ \hline

\multicolumn{2}{|l|}{\textit{Polyak--Ruppert averaging}} \\ \hline

$R_0$ &
$\displaystyle
\max\{\bE\rho_0^2,\Rx^2\}$ \\

$U_0$ &
$\displaystyle
\max\{
\bE\|\Delta_0\|^2,
C_\Delta^{(0)}+C_\Delta^{(1)}
\}$ \\

$K_Z$ &
$\displaystyle
\KM+\KMr R_0+\KMt U_0$ \\

$Q_{\beta,c}$ &
$\displaystyle
(1+2\beta)
\left(
\frac{2(\beta+1)}{ec(1-\beta)}
\right)^{\frac{\beta+1}{1-\beta}}$ \\

$C_{\bar{\Delta}}^{(0)}$ &
$\displaystyle
3\KAi^2
\bigg[
2\big(
\sqrt{\bE\|\Delta_0\|^2}
+Q_{\beta,c}\sqrt{C_\Delta^{(0)}}
\big)^2
+\bE\rho_0^2
\bigg]$ \\

$C_{\bar{\Delta}}^{(1)}$ &
$\displaystyle
3\KAi^2
\big(
18C_\Delta^{(1)}
+2\Rx^2+K_Z
\big)$ \\ \hline

\end{tabular}
\end{table}

\section{Proof of main result}
\label{sec:proof}

We begin by rewriting our update rules in Algorithm~\ref{alg: FedTD} in a form that enables our analysis. For each $i\in[N],$ let
\begin{equation}
\label{e:W.i.t + 1.defn}
        W_{t+1}^{(i)} :=  \cR_{i}(s_{t}^{i}, a_t^{i}) - (d_{i}^{\mu})^\tr \cR_{i}^{\mu} 
\end{equation}
and
\begin{align}
   M_{t+1}^{(i)} := {} &  [\cR_{i}(s_{t}^{i}, a_{t}^{i})\phi(s_{t}^{i}) - b_{i}] - [\phi(s_{t}^{i}) - v_{i}]r_{t} \nonumber  \\
   {} & \quad - [\phi(s_{t}^{i})(\phi^{\tr}(s_{t}^{i}) - \phi^{\tr}(\hat{s}_{t}^{i})) - A_{i}]\theta_{t}, \label{e:M.i.t + 1.defn}
\end{align}
where $A_{i}, b_{i},$ and $v_{i}$ are as in Section~\ref{sec:algorithm}, while $\dm_i$ and $\Rm_i$ are as in Section~\ref{sec:setup}. Further, $\forall t\geq 0,$ let
\begin{equation}
\label{defn:r_t.theta_t.bar_Delta_t}
    \rho_{t}:= r_{t} - r^{*}\text{, } \quad \Delta_{t} := \theta_{t}-\theta^{*}, \quad \text{and} \quad
    \bar{\Delta}_{t} := \bar{\theta}_{t}-\theta^{*}.
\end{equation}
Then, Algorithm~\ref{alg: FedTD}'s update rules translate to 
\begin{align}
    \rho_{t+1} = {} &  \big(1 - \frac{1}{t+1}  \big)  \rho_{t}  + \frac{1}{t+1}W_{t+1}, \label{e:rho.update.rule} \\
    \Delta_{t+1} = {} & (I - \beta_{t}A) \Delta_{t}  - \beta_{t}v\rho_{t} + \beta_{t}M_{t+1}, \label{e:Delta.update.rule} \\
    \intertext{and}
    \bar{\Delta}_{t+1} = {} & \bar{\Delta}_{t} +  \frac{1}{t+1} [\Delta_{t} - \bar{\Delta}_{t}], \label{e:Delta.bar.update.rule}
\end{align}
where 
\[
    W_{t+1}:= \frac{1}{N}\sum_{i\in [N]}W^{(i)}_{t+1} \quad \text{ and } \quad  M_{t+1}: = \frac{1}{N}\sum_{i\in[N]}M_{t+1}^{(i)}.
\]
Clearly, both $(W_{t+1})$ and $(M_{t+1})$ are Martingale-difference sequences w.r.t. the filtration $(\cF_{t}),$ where 
$\cF_{t}$ is the $\sigma$-field $\sigma(\theta_0, r_0^i,r_0, s_k^i, a_k^i, \hat{s}_k^i: i \in [N], 0 \leq k \leq t - 1).$

Next, we provide an explanation for the non-decaying constants $\Hr(\ep, \er)$ and $\Ht(\ep, \er)$ in Theorem~\ref{thm : finite-time convergence}, which we refer to the heterogeneity gaps. Applying stochastic approximation theory \cite[Chapter~2]{borkar2009} to \eqref{e:rho.update.rule}, the only potential point that $r_t$ can  converge to is $\rS.$ Similarly, from \eqref{e:Delta.update.rule} and \eqref{e:Delta.bar.update.rule}, the only point where $\Delta_t$ and, hence, $\bar{\Delta}_t$ can converge to is $\thS.$ In contrast, the standard single-agent TD(0) \cite{zhang2021finite}, if run by any agent $i$ without communication, would generate a sequence $(r_{t}, \theta_{t})$ that would converge to $(\rS_i, \thS_i).$ This difference in the limit is due to the heterogeneity in our FL setup and is the reason behind $\Hr(\ep, \er)$ and $\Ht(\ep, \er)$ in Theorem~\ref{thm : finite-time convergence}. Our first lemma formally shows that $\Ht$ (resp. $\Hr$) bounds the gap between $\thS$ (resp. $\rS$) and $\thS_i$ (resp. $\rS_i$). 

\begin{lemma}\label{thm: perturb.bound}
For each $i\in [N],$ 
\begin{equation*}
    \begin{aligned}
        |r^{*} - r^{*}_{i}| \leq {} & \sqrt{\Hr(\ep, \er)} \\
        \| \theta^{*} - \theta_{i}^{*}\| \leq {} & \sqrt{\Ht(\ep, \er)},
    \end{aligned}
\end{equation*}
where $\Hr(\ep, \er)$ and $\Ht(\ep, \er)$ are  as defined in Table \ref{tab: constants.FedTD}.
\end{lemma}
See Section~\ref{sec:technical} for a proof.

\begin{remark}
    The constants $\Hr(\ep, \er) = \Ht(\ep, \er) = 0$ when the heterogeneity parameters $\ep$ and $\er$ equal zero.
\end{remark}

The next two lemmas quantify the convergence rates of $r_t$ and $\theta_t$ to $\rS$ and $\thS,$ respectively. 
\begin{lemma}\label{thm: avg.rew.converge}
    For $T \geq 1,$
    \begin{equation}\label{e: avg.rew.converge}
        \bE\rho_{T}^{2} \leq \frac{\Rx^2}{N T} = O\left(\frac{1}{NT} \right).
    \end{equation}
\end{lemma}

See Section~\ref{sec:technical} for a proof.

\begin{remark}
    Note that $\bE \rho_T^2$ does not depend on $\rho_0.$ This is not surprising since, from \eqref{e:rho.update.rule}, we have $\rho_1 = W_1$ which clearly does not depend on $\rho_0.$ 
\end{remark}

\begin{lemma}
\label{lem:Delta.conv.rate}
Let $\beta_t=(t+1)^{-\beta}$ for some $\beta\in(0,1)$.
Then, for all $T\geq1$, we have
\begin{equation}
\label{e:Delta.conv.rate}
    \bE\|\Delta_T\|^2
    \leq C_\Delta^{(0)} \exp\left(-cT^{1-\beta}\right)
    +\frac{C_\Delta^{(1)}}{NT^\beta},
\end{equation}
where $C_\Delta^{(0)}$, $C_\Delta^{(1)}$, and $c$
are positive constants defined in
Table~\ref{tab: constants.FedTD}.
Consequently, for any fixed $N$,
\[
    \bE\|\Delta_T\|^2
    =O\left(\frac{1}{NT^\beta}\right).
\]
\end{lemma}

See Section~\ref{sec:technical} for a proof.

\begin{lemma}
\label{thm: bartheta.converge}
For all $T\geq1$, we have
\begin{equation}
\label{e:bartheta.converge}
    \bE\|\bar{\Delta}_T\|^2
    \leq
    \frac{C_{\bar{\Delta}}^{(0)}}{T^2}
    +\frac{C_{\bar{\Delta}}^{(1)}}{NT},
\end{equation}
where $C_{\bar{\Delta}}^{(0)}$ and
$C_{\bar{\Delta}}^{(1)}$ are constants
defined in Table~\ref{tab: constants.FedTD}.
\end{lemma}

\vspace{1ex}

We now prove Theorem~\ref{thm : finite-time convergence}
by combining Lemmas~\ref{thm: perturb.bound},
\ref{thm: avg.rew.converge},
\ref{lem:Delta.conv.rate}, and
\ref{thm: bartheta.converge}.

\noindent \emph{Proof of Theorem~\ref{thm : finite-time convergence}.}
Fix any $i\in[N]$ and $T\geq1$.

First, recall that
\[
    r_T-r_i^*=\rho_T+(r^*-r_i^*).
\]
Since $\bE\rho_T=0$ for $T\geq1$, we have
\begin{align*}
\bE|r_T-r_i^*|^2
&=\bE\rho_T^2+|r^*-r_i^*|^2\\
&\overset{(a)}{\leq}
\frac{\Rx^2}{NT}+\Hr(\ep,\er),
\end{align*}
where (a) follows from
Lemmas~\ref{thm: avg.rew.converge}
and \ref{thm: perturb.bound}.

Next, observe that
\[
    \bar{\theta}_T-\theta_i^*
    =\bar{\Delta}_T+(\theta^*-\theta_i^*).
\]
Consequently,
\begin{align*}
\bE\|\bar{\theta}_T-\theta_i^*\|^2
&\overset{(b)}{\leq}
2\bE\|\bar{\Delta}_T\|^2
+2\|\theta^*-\theta_i^*\|^2\\
&\overset{(c)}{\leq}
\frac{2C_{\bar{\Delta}}^{(0)}}{T^2}
+\frac{2C_{\bar{\Delta}}^{(1)}}{NT}
+2\Ht(\ep,\er).
\end{align*}
Here, (b) follows from
$\|x+y\|^2\leq2\|x\|^2+2\|y\|^2$,
while (c) follows from
Lemmas~\ref{thm: bartheta.converge}
and \ref{thm: perturb.bound}.

This completes the proof. \qed


\addtolength{\textheight}{-3cm}   

\section{Proofs of Technical Results}
\label{sec:technical}

We begin with the proof of Lemma~\ref{thm: perturb.bound}.

\noindent \emph{Proof of Lemma~\ref{thm: perturb.bound}:}
Our proof follows the approach of \cite[Theorem~1]{wang2023federated}, which establishes analogous perturbation bounds for discounted MDPs. We adapt their arguments to the average-reward setting.

First, we show that the stationary distributions $\dm_i$ and $\dm_j$ are close. For any $i, j \in [N]$ and $s \neq s' \in \cS,$ we have from Assumption~\ref{a: dynamics.hetero} that 
\begin{equation}
\label{e:Pm_i.Pm_j.Bound}
    |\Pm_i(s, s') - \Pm_j(s, s')| \leq \ep \Pm_i(s, s'). 
\end{equation}
Hence, \cite[Corollary~1]{o1993entrywise} (cf. \cite[Lemma~1]{wang2023federated}) implies  
\begin{equation}
\label{e:dmu_i.dmu_j.coordinate.bound}
    |\dm_i(s) - \dm_j(s)| \leq \dm_i(s)\ \Kd(\ep)
\end{equation}
for each $s \in \cS,$ where $\Kd(\ep)$ is as defined in Table~\ref{tab: constants.FedTD}. Therefore, 
\begin{align}
    \|\dm_i - \dm_j\|_1 & = \sum_{s \in \cS}|\dm_i(s) - \dm_j(s)| \nonumber \\
    & \overset{(a)}{\leq} \Kd(\ep) ,\nonumber
\end{align}
where (a) follows from \eqref{e:dmu_i.dmu_j.coordinate.bound}. Moreover, since the $\|\cdot\|_1$ distance between two distributions is at most $2,$ we have
\[
    \|\dm_i - \dm_j\|_1 \leq 2.
\]
Combining the two bounds gives
\begin{equation}
\label{e:dmu_i.dmu_j.Gap}
    \|\dm_i - \dm_j\|_1 \leq \min\{2, \Kd(\ep)\}.
\end{equation}

We now derive the bound on $|\rS - \rS_i|.$ For $i \neq j,$ 
\begin{align}
    |\rS_i - \rS_j| & =  |(\dm_i)^\tr \Rm_i - (\dm_j)^\tr \Rm_j| \nonumber \\
    & \overset{(a)}{\leq}\|d^\mu_i\|_1\ \|\Rm_i - \Rm_j\|_\infty + \|\Rm_j\|_\infty \ \|d^\mu_i - d^\mu_j\|_1  \nonumber \\
    & \overset{(b)}{\leq} \er + \Rx \min\{2, \Kd(\ep)\} \label{e:r.gap.interim.Bd}\\
    & \leq \sqrt{\Hr(\ep, \er)} \nonumber
\end{align}
where $\Hr$ is as in Table~\ref{tab: constants.FedTD}. The relation in (a) follows by applying triangle inequality, followed by Holder's inequality, while (b) follows because  $\|d_i^\mu\|_1 = 1,$ $\|\dm_i - \dm_j\|_1 \leq \min\{2, \Kd(\ep)\}$ from \eqref{e:dmu_i.dmu_j.Gap}, and since Assumptions~\ref{a: dynamics.hetero} and \ref{a: reward.bound} imply $\|\cR_j^{\mu}\|_\infty \leq \Rx$ and 
$|\cR_j^\mu(s) - \cR_i^\mu(s)| = \sum_{a \in \cA} \mu(a|s) |\cR_j(s, a) - \cR_i(s, a)| \leq \er$ and, hence, $\|\cR_i^\mu - \cR_j^{\mu}\|_\infty \leq \er.$ Therefore, as desired,
\[
    |\rS - \rS_i| \leq \frac{1}{N} \sum_{j \neq i} |\rS_j - \rS_i| \leq \sqrt{\Hr(\ep, \er)}.
\]

We next derive the bound on $\|\thS- \thS_i\|.$ Recall from the definitions preceding  Theorem~\ref{thm : finite-time convergence} that $A \thS = b - v \rS$ and $A_i \thS_i = b_i - v_i \rS_i.$ To establish the desired bound, we now bound $\|A - A_i\|$ and $\|(b - v \rS) - (b_i - v_i \rS_i)\|$.

We begin with a bound on $\|A-A_i\|$. By triangle inequality, 
\[
    \|A_i - A_j\| \leq \|\cT_1\| + \|\cT_2\|
\]
where 
\[
    \cT_1=\Phi^\tr(D_i^\mu-D_j^\mu)(I-\cP_j^\mu)\Phi,
    \quad
    \cT_2=\Phi^\tr D_i^\mu(\cP_j^\mu-\cP_i^\mu)\Phi.
\]
We now bound $\|\cT_1\|$ and $\|\cT_2\|$ separately. First, since $\|\Phi\| \leq 1,$ we have $\|\phi(s)\| \leq 1$ and $\|\sum_{s'} \cP_j^{\mu}(s, s') \phi(s')\| \leq 1$ for each $s \in \cS.$ Therefore, 
\begin{align*}
    \| & \cT_1\|\\
    & \leq \sum_{s} |d_i^\mu(s) - d_j^\mu(s)| \|\phi(s)\| \|\phi(s) - \sum_{s'} \cP_j^{\mu}(s, s') \phi(s')\| \\
    & \leq 2 \|d_i^\mu - d_j^\mu\|_1 \\
    & \leq 2 \min\{2, \Kd(\ep)\},
\end{align*}
where the last inequality follows from \eqref{e:dmu_i.dmu_j.Gap}. 
Similarly, 
\begin{align*}
    \|\cT_2\| & \leq \sum_{s} d_i^\mu(s) \|\phi(s)\| \sum_{s'} |\cP_j^{\mu}(s, s') - \cP_i^\mu(s, s')| \|\phi(s')\| \\
    & \overset{(a)}{\leq} \sum_{s} d_i^\mu(s) \sum_{s'} |\cP_j^\mu(s, s') - \cP_i^\mu(s, s')| \\
    & \leq \sum_{s} d_i^\mu(s) \bigg[ |\cP_j^\mu(s, s) - \cP_i^\mu(s, s)| \\
    & \hspace{2em }+ \sum_{s' \neq s}|\cP_j^\mu(s, s') - \cP_i^\mu(s, s')| \bigg] \\
    & \overset{(b)}{\leq} 2 \sum_{s} d_i^\mu(s) \sum_{s'\neq s} |\cP_j^\mu(s, s') - \cP_i^\mu(s, s')| \\
    & \overset{(c)}{\leq} 2\ep,
\end{align*}
where (a) holds since $\|\phi(s)\| \leq 1$ for all $s \in \cS,$ (b) follows because 
\begin{align*}
    |\cP_i^\mu(s, s) - \cP_j^\mu(s, s)| & = \big|\sum_{s' \neq s}(\cP_i^\mu(s, s') - \cP_j^\mu(s, s'))\big| \\
    & \leq \sum_{s' \neq s} |\cP_i^\mu(s, s') - \cP_j^\mu(s, s')|
\end{align*}
while (c) holds from Assumption~\ref{a: dynamics.hetero}. Therefore, it follows that
\begin{equation}
    \label{e:A.gap}
    \|A - A_i\| \leq \frac{1}{N} \sum_{j \neq i} \|A_j - A_i\| \leq \KA(\ep).
\end{equation}

Similarly, we obtain
\begin{align*}
\|b_i-b_j\|
& \overset{(a)}{\leq} \sum_{s\in\cS} d_i^\mu(s)
    |\Rm_i(s)-\Rm_j(s)|\|\phi(s)\|\\
&\quad + \sum_{s\in\cS}
    |d_i^\mu(s)-d_j^\mu(s)|
    |\Rm_j(s)|\|\phi(s)\|\\
&\overset{(b)}{\leq} \er+\Rx\|\dm_i-\dm_j\|_1\\
& \overset{(c)}{\leq} \er+\Rx\min\{2,\Kd(\ep)\},
\end{align*}
where (a) follows from the definitions of $b_i$ and $b_j,$ (b) holds due to Assumptions~\ref{a: dynamics.hetero} and \ref{a: reward.bound} and since  $\|\phi(s)\|\leq 1$ for all $s\in\cS$, while (c) holds from \eqref{e:dmu_i.dmu_j.Gap}.
Therefore,
\begin{equation}
\label{e:b.Bound}
\|b-b_i\|
\leq \er+\Rx\min\{2,\Kd(\ep)\}.
\end{equation}

Likewise, for any $i,j,k\in[N]$, we have
\begin{align*}
\|v_j \rS_k-v_i\rS_i\|
&\overset{(a)}{\leq}
\|v_j-v_i\|\,|\rS_k|
+\|v_i\|\,|\rS_k-\rS_i|\\
&\overset{(b)}{\leq}
2\Rx\min\{2,\Kd(\ep)\}+\er.
\end{align*}
Here, (a) follows by adding and subtracting
$v_i\rS_k$ and applying the triangle inequality, while (b) follows from
$\|v_i\|\leq 1$,
$\|v_j-v_i\|\leq\|\dm_j-\dm_i\|_1$, Assumption~\ref{a: reward.bound}, \eqref{e:r.gap.interim.Bd} and \eqref{e:dmu_i.dmu_j.Gap}. Therefore,
\begin{align}
\|v\rS-v_i\rS_i\|
&\leq \frac{1}{N^2}
\sum_{j,k\in[N]}
\|v_j\rS_k-v_i\rS_i\| \nonumber\\
&\leq 2\Rx\min\{2,\Kd(\ep)\}+\er.
\label{e:vr.bound}
\end{align}

By combining the bounds in \eqref{e:b.Bound} and \eqref{e:vr.bound}, we get
\begin{equation}
\label{e:b.gap}
    \|(b - v \rS) - (b_i - v_i \rS_i)\| \leq \Kb(\ep, \er),
\end{equation}
where $\Kb(\ep, \er)$ is as defined in Table~\ref{tab: constants.FedTD}.

Finally, we derive a bound on $\|\thS-\thS_i\|$.
Recall that $A\thS=b-v\rS$ and
$A_i\thS_i=b_i-v_i\rS_i$. Therefore,
\begin{align*}
\thS-\thS_i
={} & A^{-1}\big[(b-v\rS)-(b_i-v_i\rS_i)\big]\\
&+A^{-1}(A_i-A)\thS_i.
\end{align*}

Next, observe that
\begin{align*}
\|b_i-v_i\rS_i\|
&=\left\|
\sum_{s\in\cS}\dm_i(s)\phi(s)
\big(\Rm_i(s)-\rS_i\big)
\right\|\\
&\leq \sum_{s\in\cS}\dm_i(s)
|\Rm_i(s)-\rS_i|\\
& = \sum_{s\in\cS} \sqrt{\dm_i(s)} \sqrt{\dm_i(s)} 
|\Rm_i(s)-\rS_i| \\
&\overset{(a)}{\leq}
\sqrt{\sum_{s\in\cS}\dm_i(s)
\big(\Rm_i(s)-\rS_i\big)^2}\\
&\overset{(b)}{\leq}\Rx.
\end{align*}
Here, (a) follows from the Cauchy--Schwarz inequality,
using the fact that $\sum_s \dm_i(s)=1$.
To establish (b), recall that
$\rS_i=\sum_s\dm_i(s)\Rm_i(s)$. Therefore,
\[
    \sum_{s\in\cS}\dm_i(s) 
\big(\Rm_i(s)-\rS_i\big)^2  \leq
\sum_{s\in\cS}\dm_i(s)(\Rm_i(s))^2 \leq \Rx^2,
\]
where the last inequality follows from Assumption~\ref{a: reward.bound}. 

Consequently,
\[
\|\thS_i\|
\leq \|A_i^{-1}\|\,
\|b_i-v_i\rS_i\|
\leq \KAi\Rx.
\]
Combining these observations with
\eqref{e:A.gap} and \eqref{e:b.gap}, we obtain
\begin{align*}
\|\thS-\thS_i\|
&\leq \|A^{-1}\|
\Big[
\|(b-v\rS)-(b_i-v_i\rS_i)\|\\
&\hspace{5em}
+\|A_i-A\|\|\thS_i\|
\Big]\\
&\leq \KAi\Kb(\ep,\er)
+\KAi^2\KA(\ep)\Rx\\
&=\sqrt{\Ht(\ep,\er)},
\end{align*}
as desired. \qed

Next, we discuss the convergence rate of $r_t$ to $\rS.$


\noindent \emph{Proof of Lemma~\ref{thm: avg.rew.converge}:}
    For any $t \geq 0,$ we have that $\rho_{t} \in \cF_{t},$ while the tuple $(s^{i}_t, a^{i}_t, \hat{s}^{i}_t: i \in [N])$ and, hence, $W_{t + 1}^i$ is independent of $\cF_{t}$ for any $i \in [N].$ Furthermore, we have that $W_{t + 1}^i$ is independent of $W_{t + 1}^j$ for $i \neq j.$ Finally, we also have that $\bE[W_{t + 1}^i] = 0$ for any $i.$ Consequently,  it follows from \eqref{e:rho.update.rule} that, for any $t \geq 0,$    
    \[
        \bE \rho_{t + 1}^2 = \left(\frac{t}{t + 1}\right)^2 \bE \rho_{t}^2 + \frac{1}{N^2 (t + 1)^2} \sum_{i \in [N]} \bE (W_{t + 1}^{(i)})^2.
    \]
    Now, from \eqref{e:W.i.t + 1.defn}, we have that $W_{t+1}^{(i)}$ is a centered reward; hence, 
    \[
        \bE[(W_{t+1}^{(i)})^2]
        = \operatorname{Var}(R_i(s_t^i,a_t^i))
        \leq R_{\max}^2,
    \]
    where the last inequality follows from        
    Assumption~\ref{a: reward.bound}.
    Therefore, for any $t \geq 0,$
    \[
        \bE \rho_{t + 1}^2 \leq \left(\frac{t}{t + 1}\right)^2 \bE \rho_{t}^2 + \frac{ \Rx^2}{N (t + 1)^2}.
    \]
    A simple inductive argument now gives the desired result. \qed

To derive Lemma~\ref{lem:Delta.conv.rate}, we need a few other technical results, which we now state and prove. 

\begin{lemma}
\label{lem:M.t+1.2.Bd}
For any $t\geq 0$ and $i\in[N]$, we have
\[
    \|M_{t+1}^{(i)}\|^2
    \leq \KM+\KMr\rho_t^2+\KMt\|\Delta_t\|^2,
\]
where $\KM$, $\KMr$, and $\KMt$ are as defined
in Table~\ref{tab: constants.FedTD}.
\end{lemma}

\vspace{1ex}

\noindent \emph{Proof.}
Recall from \eqref{e:M.i.t + 1.defn} that
\[
    M_{t+1}^{(i)}
    =Z_1(t)+Z_2(t)\rho_t+Z_3(t)\Delta_t,
\]
where
\begin{align*}
Z_1(t) = {} &
    \cR_i(s_t^i,a_t^i)\phi(s_t^i)-b_i
    -[\phi(s_t^i)-v_i]\rS\\
&\quad -
    [\phi(s_t^i)(\phi^\tr(s_t^i)
    -\phi^\tr(\hat{s}_t^i))-A_i]\thS,\\
Z_2(t) = {} & -[\phi(s_t^i)-v_i],\\
Z_3(t) = {} &
    -[\phi(s_t^i)(\phi^\tr(s_t^i)
    -\phi^\tr(\hat{s}_t^i))-A_i].
\end{align*}
Now, Assumption~\ref{a: feature.matrix} implies
that $\|\phi(s)\|\leq 1$ for all $s\in\cS$.
Consequently, using the definitions of
$b_i$, $v_i$, and $A_i$, we obtain
\[
    \|b_i\|\leq\Rx,\qquad
    \|v_i\|\leq 1,\qquad
    \|A_i\|\leq 2.
\]

Further, Assumption~\ref{a: reward.bound}
implies that
$|\cR_i(s,a)|\leq\Rx$ and $|\rS|\leq\Rx$.
Therefore, by the triangle inequality,
\[
    \|Z_1(t)\|\leq 4\Rx+4\|\thS\|,
    \quad
    \|Z_2(t)\|\leq 2,
    \quad
    \|Z_3(t)\|\leq 4.
\]
Hence,
\begin{align*}
\|M_{t+1}^{(i)}\|^2
&\leq
3\|Z_1(t)\|^2
+3\|Z_2(t)\|^2\rho_t^2
+3\|Z_3(t)\|^2\|\Delta_t\|^2\\
&\overset{(a)}{\leq}
48(\Rx+\|\thS\|)^2
+12\rho_t^2+48\|\Delta_t\|^2\\
& \leq
96(\Rx^2+\|\thS\|^2)
+12\rho_t^2+48\|\Delta_t\|^2\\
&=\KM+\KMr\rho_t^2+\KMt\|\Delta_t\|^2.
\end{align*}
Here, (a) follows from the bounds on
$Z_1(t)$, $Z_2(t)$, and $Z_3(t)$.

This completes the proof. \qed

\begin{lemma}
\label{lem:Delta.one.step}
    Let $\beta_t=(t+1)^{-\beta}$ for some
    $\beta\in(0,1)$.
    Then, for all $t\geq t_\Delta$, we have
    \begin{equation}
    \label{e:Delta.one.step}
        \bE\|\Delta_{t+1}\|^2
        \leq
        \left(1-\frac{\lambda}{4}\beta_t\right)
        \bE\|\Delta_t\|^2
        +\frac{C_\Delta^{(2)}}{N}\beta_t^2,
    \end{equation}
    where $\lambda$, $t_\Delta$, and $C_\Delta^{(2)}$
    are constants defined in Table~\ref{tab: constants.FedTD}.
\end{lemma}

\noindent \emph{Proof.}
Recall from \eqref{e:Delta.update.rule} that
\[
    \Delta_{t+1}
    =(I-\beta_t A)\Delta_t-\beta_t v\rho_t
    +\beta_t M_{t+1}.
\]
Since $\rho_t$ and $\Delta_t$ are $\cF_t$-measurable
and $\bE[M_{t+1}\mid\cF_t]=0$, we have
\begin{align*}
\bE\|\Delta_{t+1}\|^2
={} & \bE\|(I-\beta_t A)\Delta_t
         -\beta_t v\rho_t\|^2
    +\beta_t^2\bE\|M_{t+1}\|^2\\
={} & \bE\|(I-\beta_t A)\Delta_t\|^2\ -2\beta_t v^\tr(I-\beta_t A)
    \bE[\rho_t\Delta_t]\\
&+\beta_t^2\|v\|^2\bE\rho_t^2
    +\beta_t^2\bE\|M_{t+1}\|^2.
\end{align*}

Next, observe that
\begin{align*}
\bE\|(I-\beta_t A)\Delta_t\|^2
&\overset{(a)}{\leq}
\|I-\beta_t A\|^2\bE\|\Delta_t\|^2\\
&\overset{(b)}{\leq}
\left[1-\beta_t\lambda_{\min}(A+A^\tr)
+\beta_t^2\|A\|^2\right]
\bE\|\Delta_t\|^2\\
&\overset{(c)}{\leq}
(1-\lambda\beta_t+4\beta_t^2)
\bE\|\Delta_t\|^2.
\end{align*}
Here, (a) follows from the definition of the
operator norm, while (b) follows by expanding
\[
(I-\beta_t A)^\tr(I-\beta_t A)
=I-\beta_t(A+A^\tr)+\beta_t^2A^\tr A.
\]
Finally, (c) holds since
$\lambda<\lambda_{\min}(A+A^\tr)$
and $\|A\|\leq 2$.

Next, consider the cross term. Since $\|v\|\leq 1$,
$\|A\|\leq 2$, and $\beta_t\leq 1$, we have
$\|I-\beta_t A\|\leq 3$. Therefore, for $t\geq 1$,
\begin{align*}
-2\beta_t & v^\tr(I-\beta_t A) 
    \bE[\rho_t\Delta_t]\\
&\overset{(a)}{\leq}
    6\beta_t\bE[|\rho_t|\|\Delta_t\|]\\
&\overset{(b)}{\leq}
    \frac{\lambda}{2}\beta_t
    \bE\|\Delta_t\|^2
    +\frac{18}{\lambda}\beta_t
    \bE\rho_t^2\\
&\overset{(c)}{\leq}
    \frac{\lambda}{2}\beta_t
    \bE\|\Delta_t\|^2
    +\frac{36\Rx^2}{\lambda N}\beta_t^2.
\end{align*}
Here, (a) follows from the Cauchy--Schwarz inequality
and the bound $\|I-\beta_t A\|\leq 3$, (b) follows from Young's inequality, while (c) follows from
Lemma~\ref{thm: avg.rew.converge} and
$1/t\leq 2\beta_t$ for $t\geq 1$.

Next, consider the remaining two terms. For $t\geq 1$,
\begin{align*}
&\beta_t^2\|v\|^2\bE\rho_t^2
+\beta_t^2\bE\|M_{t+1}\|^2\\
&\overset{(a)}{\leq}
\beta_t^2\bE\rho_t^2
+\frac{\beta_t^2}{N}
\left[
\KM+\KMr\bE\rho_t^2
+\KMt\bE\|\Delta_t\|^2
\right]\\
&\overset{(b)}{\leq}
\frac{\beta_t^2}{N}
\left[
\KM+(1+\KMr)\Rx^2
+\KMt\bE\|\Delta_t\|^2
\right].
\end{align*}
Here, (a) follows from $\|v\|\leq 1$ and
Lemma~\ref{lem:M.t+1.2.Bd}, together with
the conditional independence and zero-mean
properties of the noise terms across agents.
Inequality (b) follows from
Lemma~\ref{thm: avg.rew.converge}, which gives
$\bE\rho_t^2\leq\Rx^2/(Nt)$, and the fact
that $t,N\geq 1$.

Combining the preceding bounds, we obtain, for $t\geq 1$,
\begin{multline}
\label{e:Delta.t.gen.Bd}
\bE\|\Delta_{t+1}\|^2
\leq
\left[
1-\frac{\lambda}{2}\beta_t
+\left(4+\frac{\KMt}{N}\right)\beta_t^2
\right]\bE\|\Delta_t\|^2\\
+\frac{C_\Delta^{(2)}}{N}\beta_t^2,
\end{multline}
where
\[
C_\Delta^{(2)}
:=\KM+\left(1+\KMr+\frac{36}{\lambda}\right)\Rx^2.
\]
Now, since $\beta_t=(t+1)^{-\beta}$ and $N\geq1$,
for all $t\geq t_\Delta$, we have
\[
\left(4+\frac{\KMt}{N}\right)\beta_t
\leq (4+\KMt)\beta_t
\leq\frac{\lambda}{4},
\]
where $t_\Delta$ is as defined in
Table~\ref{tab: constants.FedTD}. Consequently,
\[
\bE\|\Delta_{t+1}\|^2
\leq
\left(1-\frac{\lambda}{4}\beta_t\right)
\bE\|\Delta_t\|^2
+\frac{C_\Delta^{(2)}}{N}\beta_t^2,
\]
for all $t\geq t_\Delta$, as desired. \qed

\vspace{1ex}

\noindent \emph{Proof of Lemma~\ref{lem:Delta.conv.rate}}   
Let $u_t:=\bE\|\Delta_t\|^2$ and
$a:=\lambda/4$.
By Lemma~\ref{lem:Delta.one.step}, for all
$t\geq t_\Delta$, we have
\begin{equation}
\label{e:Delta.scalar.recursion}
u_{t+1}
\leq (1-a\beta_t)u_t
+\frac{C_\Delta^{(2)}}{N}\beta_t^2.
\end{equation}
Next, define
\[
t_*:=\max\left\{
t_\Delta,\,
\left\lceil
\left(\frac{2\beta}{a}\right)^{1/(1-\beta)}
\right\rceil
\right\}.
\]
Since $\beta_t=(t+1)^{-\beta}$, the mean value
theorem implies that, for $t\geq t_*$,
\begin{align*}
\beta_t-\beta_{t+1}
&\leq \beta(t+1)^{-\beta-1}\\
&\leq \frac{a}{2}\beta_t^2.
\end{align*}
Consequently,
\begin{equation}
\label{e:stepsize.comparison}
\beta_{t+1}
\geq \beta_t-\frac{a}{2}\beta_t^2,
\qquad t\geq t_*.
\end{equation}

Let
\[
    D:=\frac{2C_\Delta^{(2)}}{a}
    =\frac{8C_\Delta^{(2)}}{\lambda}.
\]
We claim that, for all $T\geq t_*$,
\begin{equation}
\label{e:Delta.comparison.bound}
    u_T
    \leq
    u_{t_*}\prod_{t=t_*}^{T-1}(1-a\beta_t)
    +\frac{D\beta_T}{N},
\end{equation}
where an empty product is taken to be one.
We prove this by induction on $T$.
The claim is immediate for $T=t_*$.
Now, suppose it holds for some $T\geq t_*$.
Then, from \eqref{e:Delta.scalar.recursion},
\begin{align*}
u_{T+1}
&\leq (1-a\beta_T)u_T
    +\frac{C_\Delta^{(2)}}{N}\beta_T^2\\
&\leq
u_{t_*}\prod_{t=t_*}^{T}(1-a\beta_t)\\
&\quad+
\frac{D}{N}
\left[
\beta_T-a\beta_T^2
+\frac{C_\Delta^{(2)}}{D}\beta_T^2
\right]\\
&\overset{(a)}{=}
u_{t_*}\prod_{t=t_*}^{T}(1-a\beta_t)
+\frac{D}{N}
\left[\beta_T-\frac{a}{2}\beta_T^2\right]\\
&\overset{(b)}{\leq}
u_{t_*}\prod_{t=t_*}^{T}(1-a\beta_t)
+\frac{D\beta_{T+1}}{N}.
\end{align*}
Here, (a) follows from the choice
$D=2C_\Delta^{(2)}/a$, while (b) follows from
\eqref{e:stepsize.comparison}.
This completes the induction.

Next, for any $T\geq t_*$, we have
\begin{align}
\prod_{t=t_*}^{T-1} & (1-a\beta_t) \nonumber \\
&\overset{(a)}{\leq}
\exp\left(-a\sum_{t=t_*}^{T-1}\beta_t\right) \nonumber \\
&\overset{(b)}{\leq}
\exp\left[
-\frac{a}{1-\beta}
\left[(T+1)^{1-\beta}
-(t_*+1)^{1-\beta}\right]
\right] \nonumber \\
&\overset{(c)}{\leq}
\exp\left(c(t_*+1)^{1-\beta}\right)
\exp\left(-cT^{1-\beta}\right), \label{e:prod.Bd}
\end{align}
where
\[
    c:=\frac{a}{1-\beta}
    =\frac{\lambda}{4(1-\beta)}.
\]
Here, (a) follows from $1-x\leq e^{-x}$,
(b) follows from $\beta_t=(t+1)^{-\beta}$
and the integral bound
\[
    \sum_{t=t_*}^{T-1}\beta_t
    \geq \int_{t_*+1}^{T+1}x^{-\beta}\,dx,
\]
while (c) follows from $(T+1)^{1-\beta}
\geq T^{1-\beta}$.

We next bound $u_{t_*}$. Recall that
$u_t=\bE\|\Delta_t\|^2$. Since $\beta_0=1$,
$\|I-A\|\leq 3$, and $\|v\|\leq 1$,
the update rule in \eqref{e:Delta.update.rule} and the martingale-difference
property give
\begin{align*}
u_1
&=\bE\|(I-A)\Delta_0-v\rho_0\|^2
  +\bE\|M_1\|^2\\
&\overset{(a)}{\leq} 18u_0+2\bE\rho_0^2\\
&\quad+\frac{1}{N}
\left[\KM+\KMr\bE\rho_0^2+\KMt u_0\right]\\
&\overset{(b)}{\leq} (18+\KMt)u_0
 +(2+\KMr)\bE\rho_0^2+\KM\\
&=:U_1,
\end{align*}
where (a) follows from Lemma~\ref{lem:M.t+1.2.Bd}, while (b) follows since $N \geq 1.$

Furthermore, by using $N \geq 1$ and $\beta_t^2 \leq 1$ in our mean-square bound in \eqref{e:Delta.t.gen.Bd}, it follows 
that, for every $t\geq1$,
\[
    u_{t+1}\leq Bu_t+C_\Delta^{(2)},
    \qquad B:=5+\KMt.
\]
Applying this inequality recursively, we obtain
\begin{align}
u_{t_*}
&\leq B^{t_*-1}U_1
+C_\Delta^{(2)}
\sum_{k=0}^{t_*-2}B^k \nonumber \\
&\leq B^{t_*-1}
\left(U_1+\frac{C_\Delta^{(2)}}{B-1}\right) \nonumber \\
&=:U_*. \label{e:u.t.star.Bd}
\end{align}

Combining \eqref{e:Delta.comparison.bound} with \eqref{e:prod.Bd} and \eqref{e:u.t.star.Bd}, we obtain, for all
$T\geq t_*$,
\begin{align*}
u_T
&\leq U_*
\exp\left(c(t_*+1)^{1-\beta}\right)
\exp\left(-cT^{1-\beta}\right)
+\frac{D}{NT^\beta}\\
&=C_\Delta^{(0)}
\exp\left(-cT^{1-\beta}\right)
+\frac{C_\Delta^{(1)}}{NT^\beta},
\end{align*}
where
\[
C_\Delta^{(0)}
:=U_*\exp\left(c(t_*+1)^{1-\beta}\right),
\qquad
C_\Delta^{(1)}:=D.
\]
Moreover, the recursive bound used to define $U_*$
also implies that $u_T\leq U_*$ for
$1\leq T<t_*$. Consequently,
\[
u_T\leq
C_\Delta^{(0)}\exp(-cT^{1-\beta})
+\frac{C_\Delta^{(1)}}{NT^\beta}
\]
holds for every $T\geq1$.
Since the exponential term decays faster than
$T^{-\beta}$, the desired asymptotic bound
follows for any fixed $N$. \qed

\vspace{1ex}

\noindent \emph{Proof of Lemma~\ref{thm: bartheta.converge}:}
Recall from \eqref{e:Delta.update.rule} that
\[
A\Delta_t
=\frac{\Delta_t-\Delta_{t+1}}{\beta_t}
-v\rho_t+M_{t+1}.
\]
Summing over $t=0,\ldots,T-1$, we obtain
\[
A\bar{\Delta}_T
=\frac{1}{T}(B_T-vE_T+Z_T),
\]
where
\[
B_T:=\sum_{t=0}^{T-1}
\frac{\Delta_t-\Delta_{t+1}}{\beta_t},
\quad
E_T:=\sum_{t=0}^{T-1}\rho_t,
\quad
Z_T:=\sum_{t=0}^{T-1}M_{t+1}.
\]
Since $\|A^{-1}\|\leq\KAi$ and $\|v\|\leq1$,
\begin{equation}
\label{e:PR.master.bound}
\bE\|\bar{\Delta}_T\|^2
\leq\frac{3\KAi^2}{T^2}
\left[
\bE\|B_T\|^2+\bE E_T^2
+\bE\|Z_T\|^2
\right].
\end{equation}
We now bound the three terms separately.

First, the recursion for $\rho_t$ gives
\[
E_T=\rho_0+
\sum_{k=1}^{T-1}
\left(\sum_{t=k}^{T-1}\frac1t\right)W_k.
\]
Using the martingale-difference property,
$\bE W_k^2\leq\Rx^2/N$, and the identity
\[
\sum_{k=1}^n
\left(\sum_{t=k}^n\frac1t\right)^2
=2n-\sum_{t=1}^n\frac1t
\leq2n,\qquad n\geq1,
\]
we obtain
\begin{equation}
\label{e:PR.E.bound}
\bE E_T^2
\leq \bE\rho_0^2+\frac{2\Rx^2T}{N}.
\end{equation}
For $T=1$, the same bound follows directly
from $E_1=\rho_0$.

Next, the martingale-difference property gives
\[
\bE\|Z_T\|^2
=\sum_{t=0}^{T-1}\bE\|M_{t+1}\|^2.
\]
Moreover, since
$M_{t+1}=N^{-1}\sum_{i=1}^N M_{t+1}^{(i)}$,
conditional independence and the zero-mean
property of the agents' noise terms imply
\[
\bE\|M_{t+1}\|^2
=\frac{1}{N^2}
\sum_{i=1}^N\bE\|M_{t+1}^{(i)}\|^2.
\]
Consequently, Lemma~\ref{lem:M.t+1.2.Bd}
yields
\[
\bE\|M_{t+1}\|^2
\leq\frac{1}{N}
\left[
\KM+\KMr\bE\rho_t^2
+\KMt\bE\|\Delta_t\|^2
\right].
\]
Using Lemmas~\ref{thm: avg.rew.converge}
and \ref{lem:Delta.conv.rate}, we therefore obtain
\begin{equation}
\label{e:PR.Z.bound}
\bE\|Z_T\|^2
\leq\frac{K_ZT}{N},
\end{equation}
where $K_Z$, $R_0$, and $U_0$ are defined
in Table~\ref{tab: constants.FedTD}.

Finally, we bound $\bE\|B_T\|^2$.
Since $\beta_t=(t+1)^{-\beta}$,
summation by parts gives
\[
B_T=\Delta_0-T^\beta\Delta_T+
\sum_{t=1}^{T-1}
\left[(t+1)^\beta-t^\beta\right]\Delta_t.
\]
By Minkowski's inequality and
Lemma~\ref{lem:Delta.conv.rate},
\begin{align*}
\left(\bE\|B_T\|^2\right)^{1/2}
\leq {}&
\sqrt{\bE\|\Delta_0\|^2}\\
&+\sqrt{C_\Delta^{(0)}}\,Q_{\beta,c}\\
&+3\sqrt{\frac{C_\Delta^{(1)}}{N}}T^{\beta/2},
\end{align*}
where $Q_{\beta,c}$ is defined in
Table~\ref{tab: constants.FedTD}.

To establish the preceding bound, observe that
\[
\sup_{x>0}x^{\beta+1}e^{-cx^{1-\beta}/2}
=
\left(
\frac{2(\beta+1)}{ec(1-\beta)}
\right)^{\frac{\beta+1}{1-\beta}}
=:q_{\beta,c}.
\]
Since $(t+1)^\beta-t^\beta\leq
\beta t^{\beta-1}$ for $t\geq1$, we obtain
\begin{align*}
&T^\beta e^{-cT^{1-\beta}/2}
+\sum_{t=1}^{T-1}
\big[(t+1)^\beta-t^\beta\big]
e^{-ct^{1-\beta}/2}\\
&\quad\leq
q_{\beta,c}
\left(
\frac{1}{T}
+\beta\sum_{t=1}^{T-1}\frac{1}{t^2}
\right)\\
&\quad\leq(1+2\beta)q_{\beta,c}
=Q_{\beta,c}.
\end{align*}
Similarly, using
\[
\sum_{t=1}^{T-1}t^{\beta/2-1}
\leq
1+\frac{2}{\beta}
\big[(T-1)^{\beta/2}-1\big],
\qquad T\geq2,
\]
we obtain
\begin{align*}
&T^{\beta/2}
+\sum_{t=1}^{T-1}
\big[(t+1)^\beta-t^\beta\big]t^{-\beta/2}\\
&\quad\leq
T^{\beta/2}
+\beta\sum_{t=1}^{T-1}t^{\beta/2-1}\\
&\quad\leq3T^{\beta/2}.
\end{align*}
The latter inequality also holds for $T=1$.

Consequently, using $(a+b)^2\leq2(a^2+b^2)$,
\begin{align}
\bE\|B_T\|^2
\leq {}&
2\left(
\sqrt{\bE\|\Delta_0\|^2}
+Q_{\beta,c}\sqrt{C_\Delta^{(0)}}
\right)^2\nonumber\\
&+\frac{18C_\Delta^{(1)}}{N}T^\beta.
\label{e:PR.B.bound}
\end{align}
Combining \eqref{e:PR.master.bound},
\eqref{e:PR.E.bound}, \eqref{e:PR.Z.bound},
and \eqref{e:PR.B.bound}, and using
$T^\beta\leq T$ for $\beta\in(0,1)$, we obtain
\[
\bE\|\bar{\Delta}_T\|^2
\leq
\frac{C_{\bar{\Delta}}^{(0)}}{T^2}
+\frac{C_{\bar{\Delta}}^{(1)}}{NT},
\]
where
\begin{align*}
C_{\bar{\Delta}}^{(0)}
:={}&3\KAi^2
\bigg[
2\left(
\sqrt{\bE\|\Delta_0\|^2}
+Q_{\beta,c}\sqrt{C_\Delta^{(0)}}
\right)^2
+\bE\rho_0^2
\bigg],\\
C_{\bar{\Delta}}^{(1)}
:={}&3\KAi^2
\left(
18C_\Delta^{(1)}
+2\Rx^2+K_Z
\right).
\end{align*}
This completes the proof. \qed

\begin{figure*}
    \begin{subfigure}[b]{1\columnwidth}
        \includegraphics[width=\textwidth]{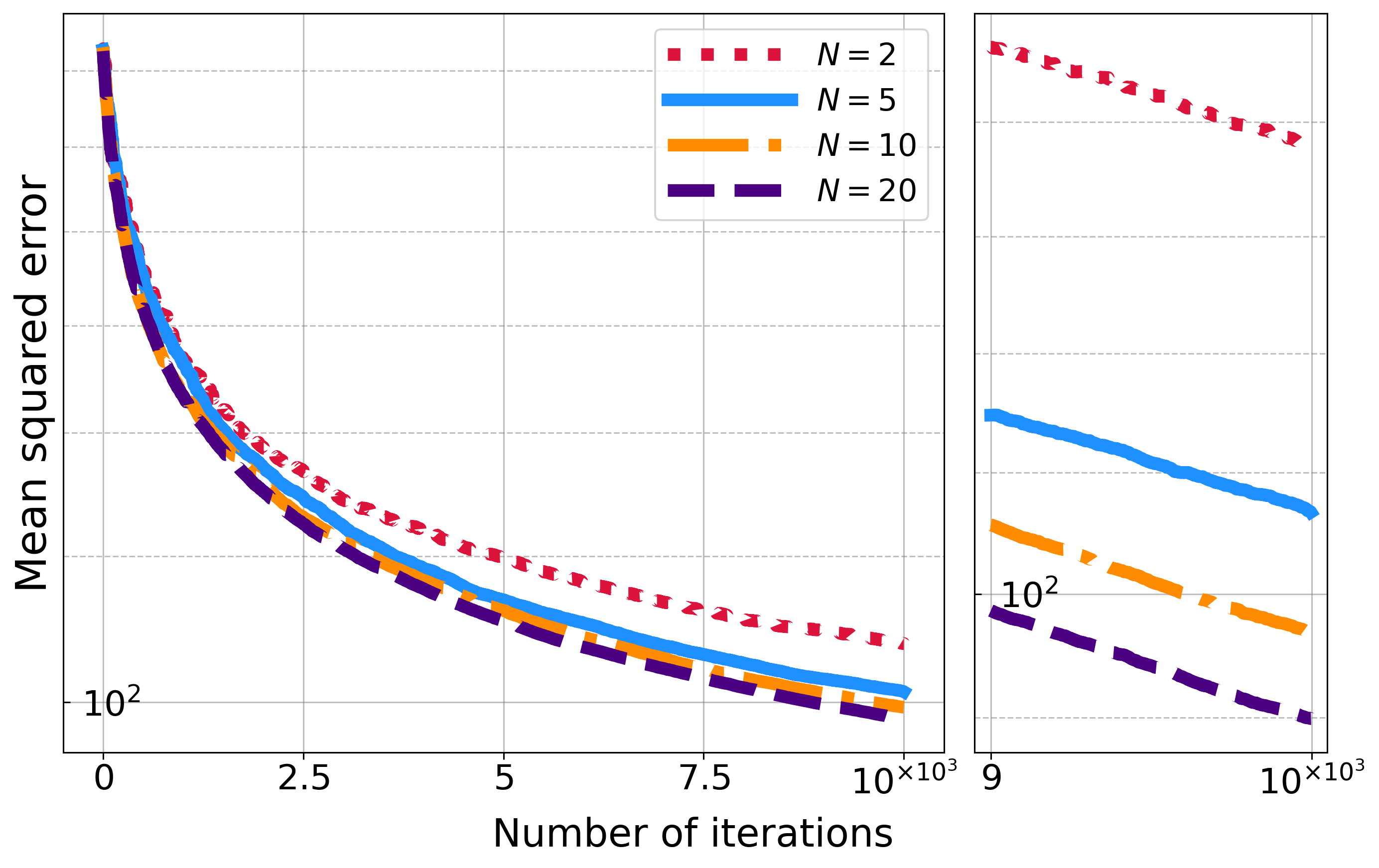}
        \caption{Proposed algorithm, $\epsilon_p=\epsilon_r=0.1$}
        \label{fig:quad_b}
        \label{fig:quad_a}
    \end{subfigure}
    \hfill
    \begin{subfigure}[b]{1\columnwidth}
        \includegraphics[width=\textwidth]{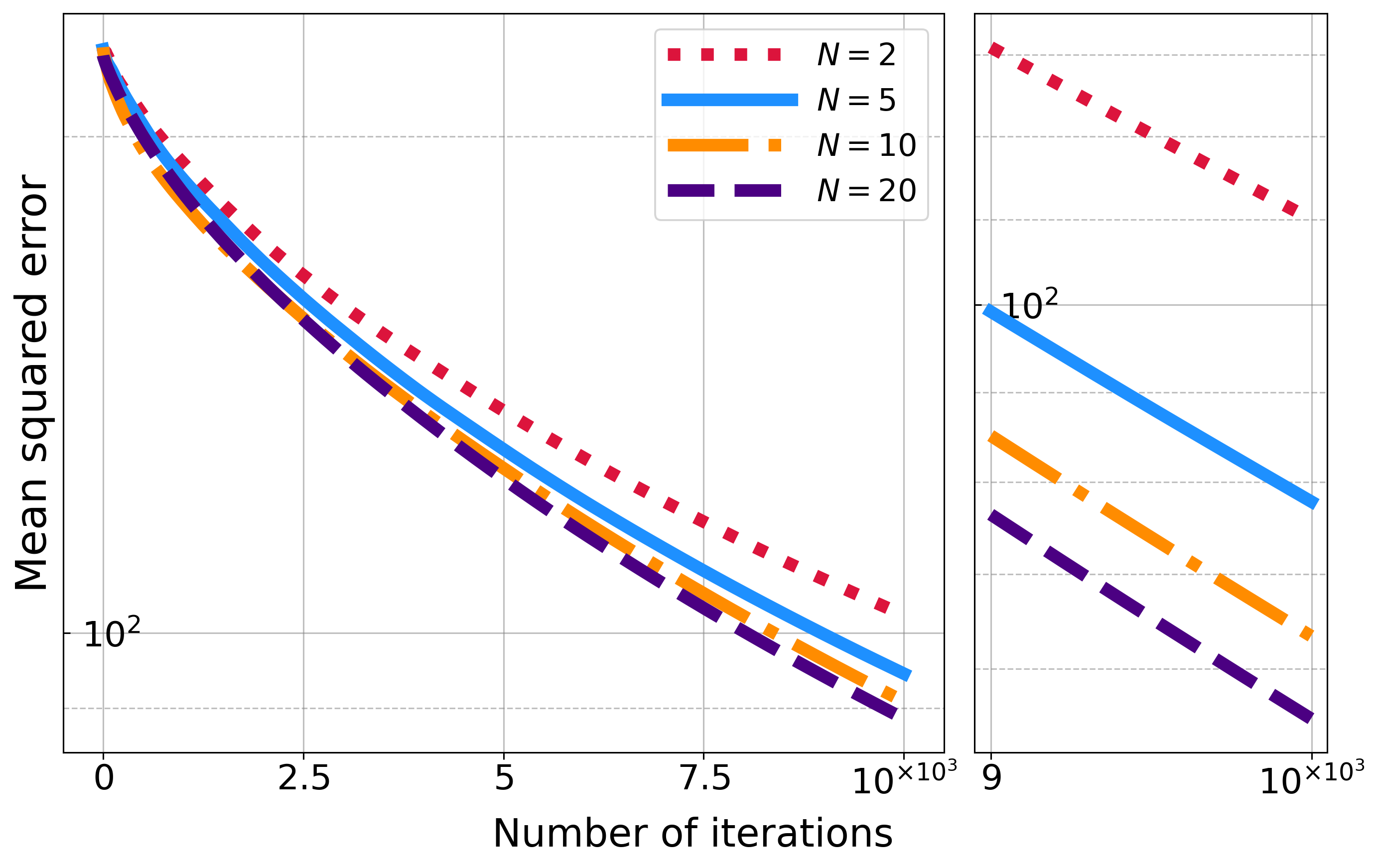}
        \caption{Vanilla algorithm, $\epsilon_p=\epsilon_r=0.1$}
    \end{subfigure}

    \begin{subfigure}[b]{1\columnwidth}
        \includegraphics[width=\textwidth]{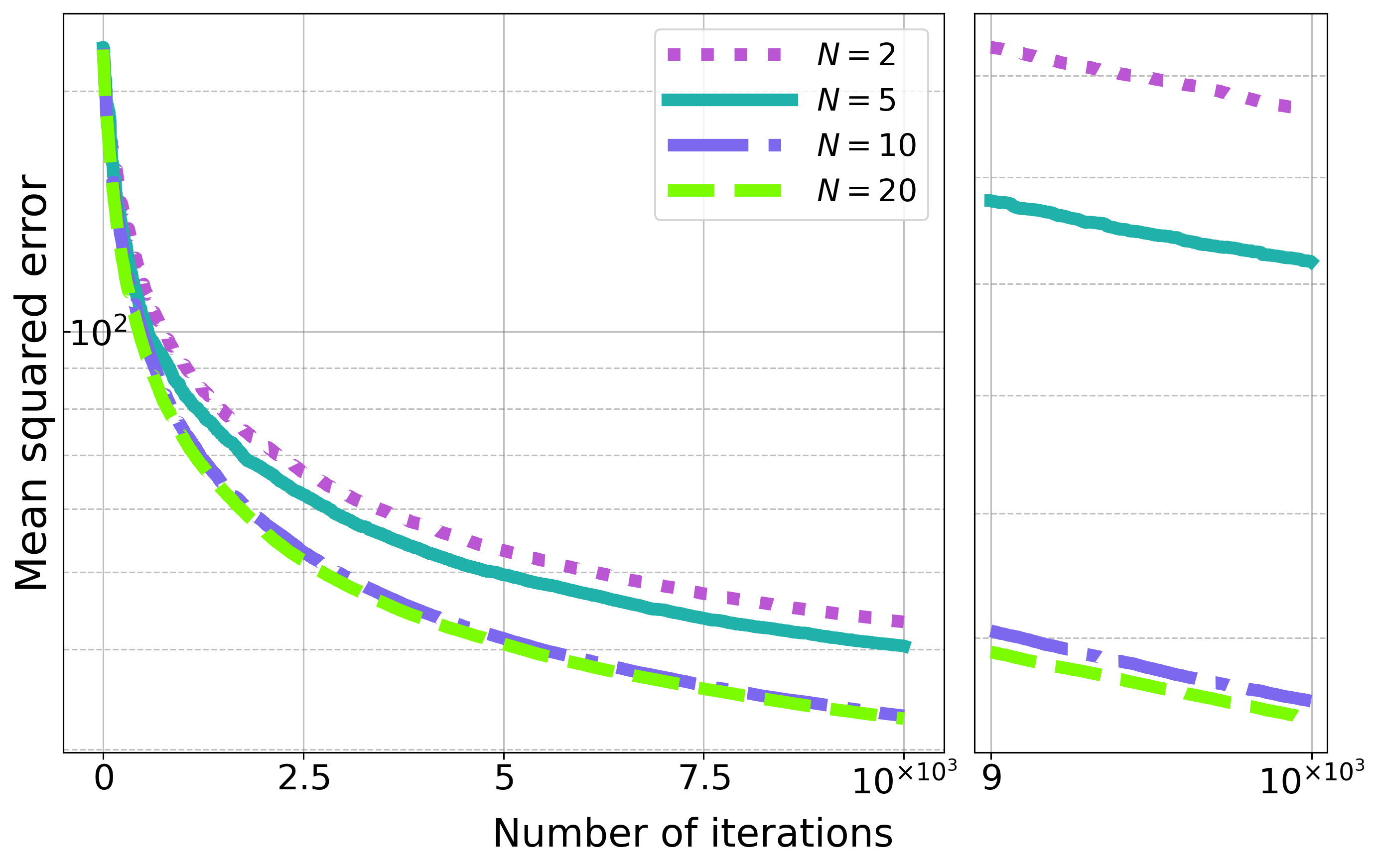}
        \caption{Proposed algorithm, $\epsilon_p=\epsilon_r=0.7$}
        \label{fig:quad_c}
    \end{subfigure}
    \hfill
    \begin{subfigure}[b]{1\columnwidth}
        \includegraphics[width=\textwidth]{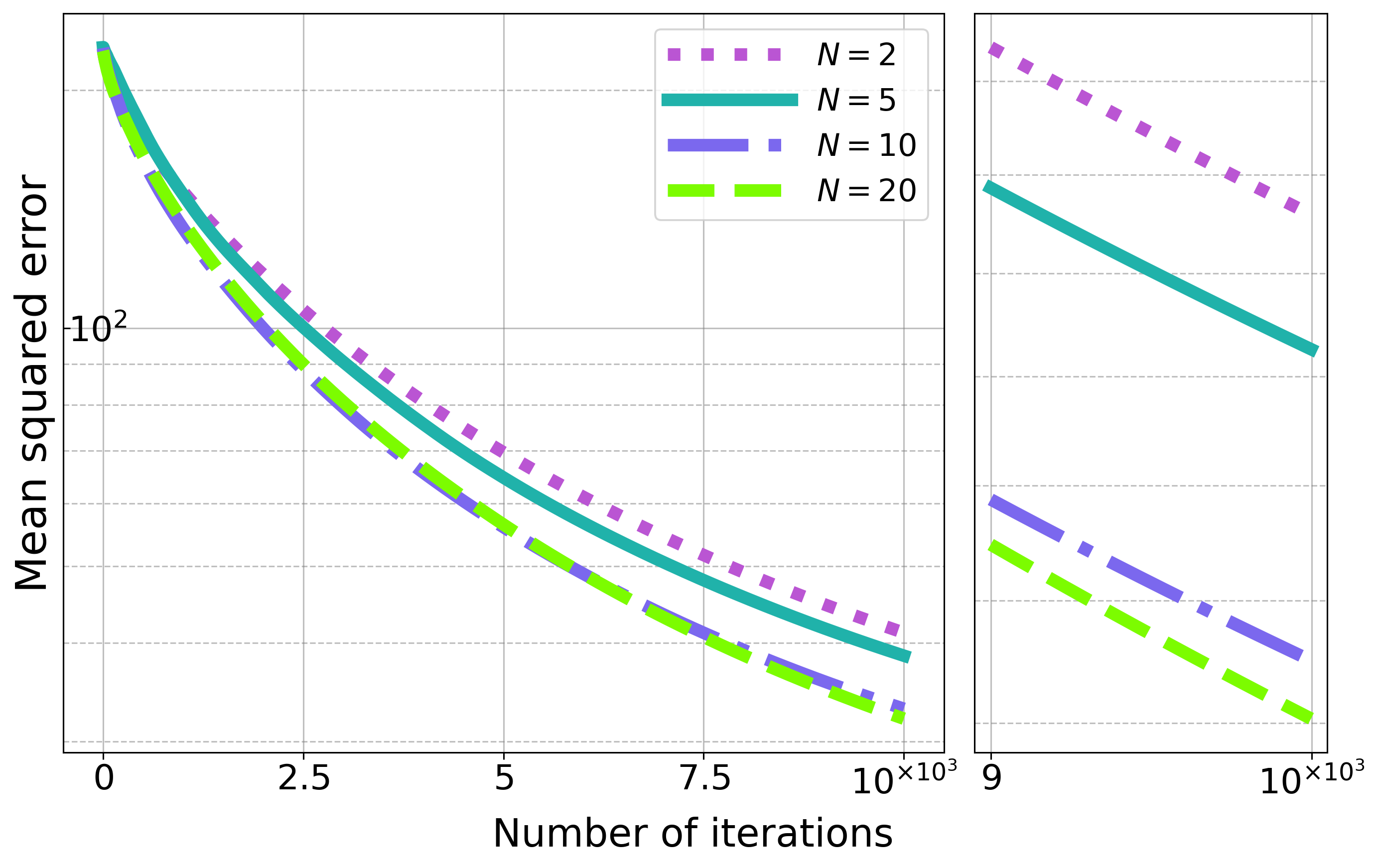}
        \caption{Vanilla algorithm, $\epsilon_p=\epsilon_r=0.7$}
        \label{fig:quad_d}
    \end{subfigure}
    \caption{Comparison of our proposed two-timescale federated TD algorithm (left column) with the vanilla single-timescale federated TD algorithm described in Section~\ref{sec:algorithm} (right column). The top row corresponds to \(\epsilon_p=\epsilon_r=0.1\), while the bottom row corresponds to \(\epsilon_p=\epsilon_r=0.7\). We consider \(N\in\{2,5,10,20\}\) agents. The horizontal axis shows the number of iterations, while the vertical axis shows the mean-squared error relative to \(\theta_1^*\), computed using the Polyak–Ruppert averaged iterate for our algorithm and the raw iterate for the vanilla algorithm. Each curve is averaged over 10 independent runs, and the final 1000 iterations are magnified in the inset plots. Our algorithm uses \(\beta=0.3\). Both algorithms exhibit lower errors as \(N\) increases, with comparable performance despite our algorithm not requiring model-dependent stepsizes.}
    \label{fig:quad}
\end{figure*}

\section{Experiments}
\label{sec: exp}

In this section, we evaluate our proposed algorithm against the vanilla single-timescale federated TD algorithm  obtained by combining the approaches of Wang et al.~\cite{wang2023federated} and Zhang et al.~\cite{zhang2021finite}, as described in Section~\ref{sec:algorithm}. Our experiments address three questions: (i) How does performance vary with the number of agents and the level of heterogeneity? (ii) How does the choice of stepsize affect performance? (iii) How does heterogeneity affect long-term convergence? The results show that our algorithm benefits from increasing the number of agents and achieves performance comparable to, and sometimes better than, the vanilla baseline, without requiring knowledge of unknown model parameters for stepsize selection. We also observe a residual error due to heterogeneity, consistent with our theoretical findings.

\subsection{Experimental Setup}
    We consider the heterogeneous federated policy evaluation setting described in Section~\ref{sec:setup}. All agents share the same state and action spaces, with \(|\mathcal S|=|\mathcal A|=100\), and use a common feature matrix \(\Phi\in\mathbb R^{100\times20}\). We consider four choices for the number of agents: \(N\in\{2,5,10,20\}\) and two levels of heterogeneity: \(\epsilon_p=\epsilon_r=0.1\) and \(\epsilon_p=\epsilon_r=0.7\). The reward and transition functions of the first agent's MDP, \(\mathcal M_1\), are generated randomly and kept fixed across experiments. For each choice of the heterogeneity parameters, the reward and transition functions of the remaining agents are generated randomly, subject to the bounds in Assumption~\ref{a: dynamics.hetero}.

    We compare our proposed Algorithm~\ref{alg: FedTD} with the vanilla single-timescale algorithm described in Section~\ref{sec:algorithm}. Our algorithm uses stepsizes \(1/(t+1)\) and \(1/(t+1)^\beta\), respectively, for updating the average-reward estimate $r_t^i$ and the value-function parameters $\theta_t$, where \(\beta\in(0,1)\). The vanilla algorithm uses a stepsizes of the form $c_\alpha \beta_t$ and \(\beta_t=c_1/(c_2+t)\) for making similar updates, where the constants \(c_1\), \(c_2\), and \(c_\alpha\) must satisfy conditions specified in \cite{zhang2021finite} that depend on unknown parameters (see Section~\ref{sec:algorithm}). To compare their performance, we measure the squared error of the Polyak–Ruppert averaged iterate, \(\|\bar\theta_t-\theta_1^*\|^2\), for our algorithm and that of the raw iterate, \(\|\theta_t-\theta_1^*\|^2\), for the vanilla algorithm, where \(\theta_1^*\) is the solution corresponding to \(\mathcal M_1\). The reported errors are averaged over multiple runs.

\begin{figure}[htb]
    \centering
    \includegraphics[width=0.47\textwidth]{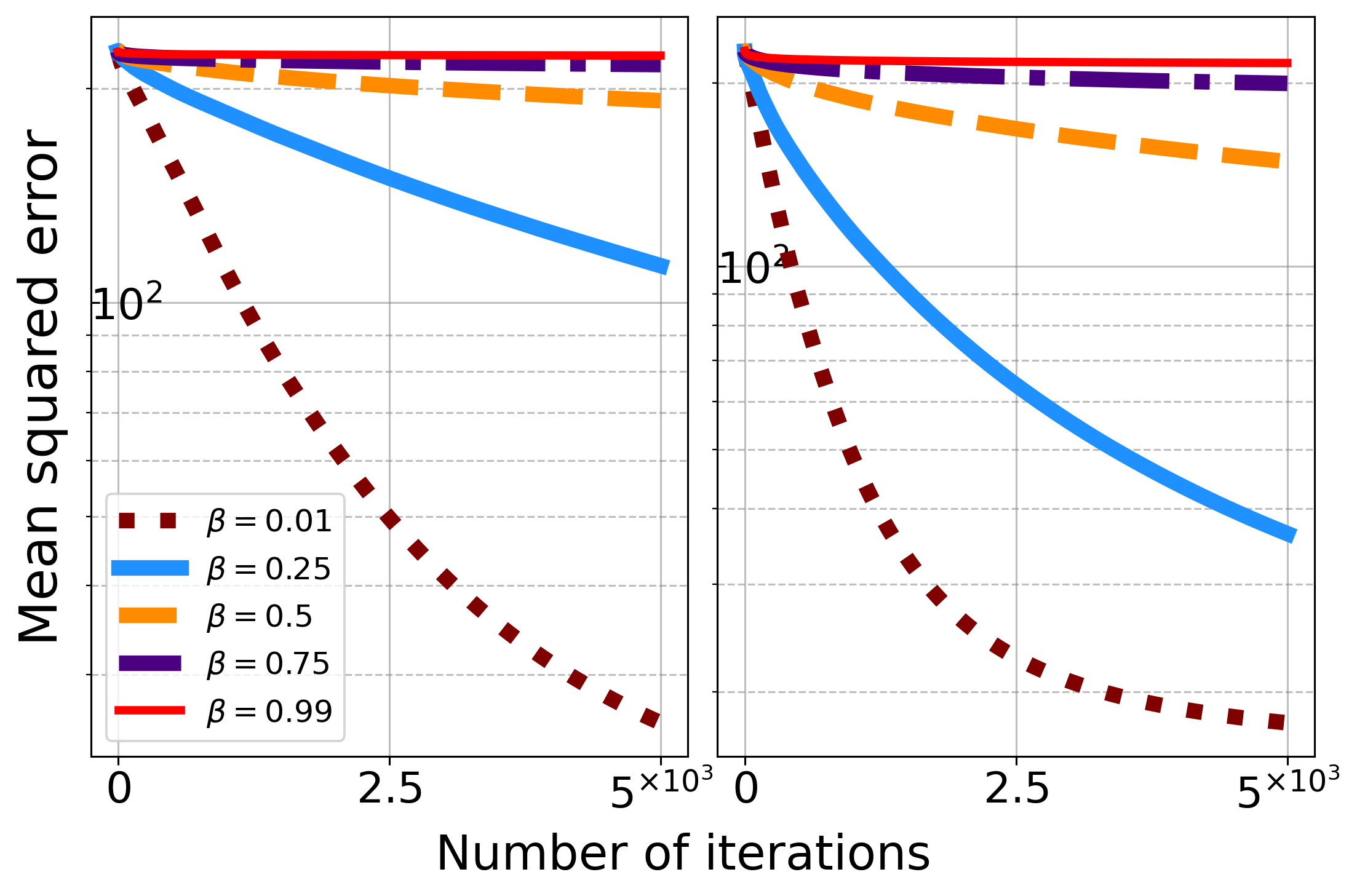}
    \caption{Effect of the stepsize exponent \(\beta\) on the performance of Algorithm~\ref{alg: FedTD} for \(N=5\) agents. We vary \(\beta\) from \(0.01\) to \(0.99\). The left panel corresponds to \(\epsilon_p=\epsilon_r=0.1\), while the right panel corresponds to \(\epsilon_p=\epsilon_r=0.7\). Each curve shows the mean-squared error averaged over 10 independent runs of 5000 iterations each. Smaller values of \(\beta\) generally lead to faster error reduction in both settings.}
    \label{fig:beta}
\end{figure}

\begin{figure}[htb]
    \centering
    \includegraphics[width=0.67\textwidth]{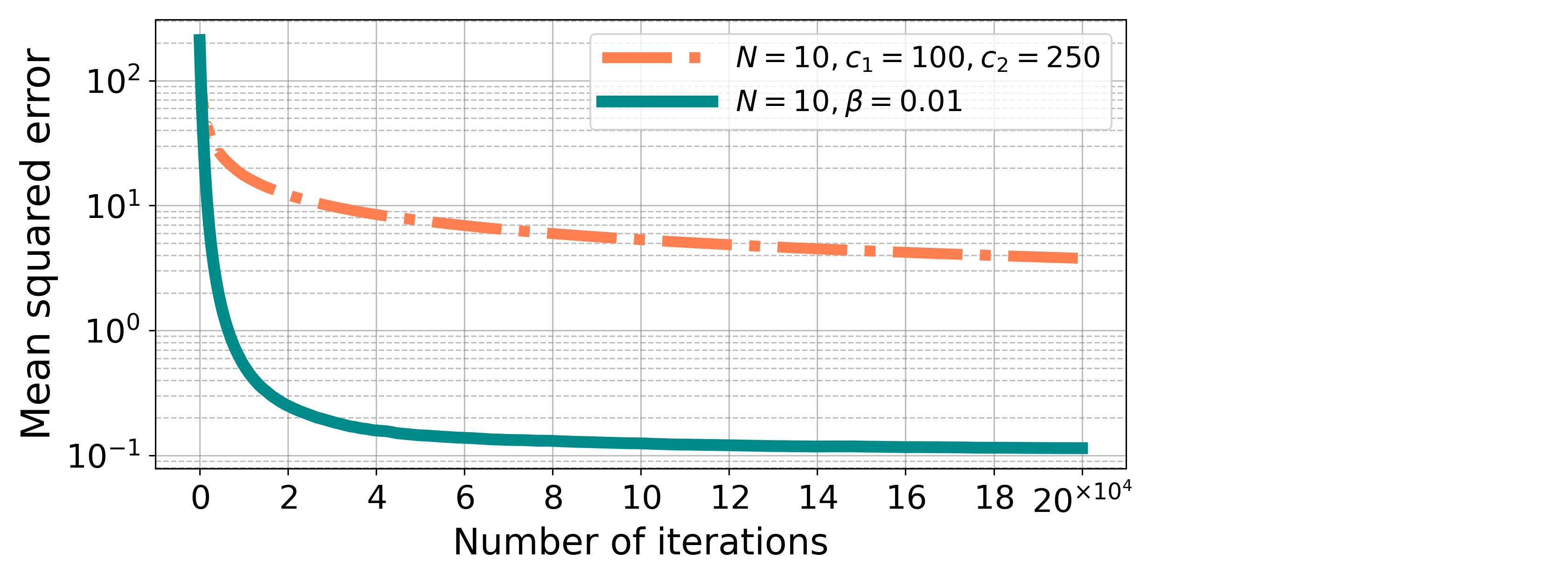}
    \caption{Long-term performance of our proposed Algorithm~\ref{alg: FedTD} (solid line) and the vanilla single-timescale federated TD algorithm (dotted line) for \(N=10\) agents with \(\epsilon_p=\epsilon_r=0.7\). Each curve is averaged over 5 independent runs of 200,000 iterations. The errors eventually level off, consistent with the residual error due to heterogeneity predicted by our theoretical results. Our algorithm achieves a substantially lower error than the vanilla baseline in this experiment.}
    \label{fig:heavyrun}
\end{figure}

\subsection{Effect of the Number of Agents and Heterogeneity}

Figure~\ref{fig:quad} compares the performance of our proposed algorithm with that of the vanilla single-timescale algorithm for different numbers of agents and levels of heterogeneity. In both heterogeneity settings, increasing the number of agents reduces the mean-squared error. Also, our algorithm achieves performance comparable to the vanilla algorithm, despite using stepsizes that do not depend on unknown parameters.

These observations are consistent with Theorem~\ref{thm : finite-time convergence}. In particular, increasing \(N\) reduces the stochastic estimation error, while the error due to heterogeneity need not vanish. Thus, the experiments illustrate the benefits of federated learning while highlighting the practical advantage of our model-independent stepsize selection.

\subsection{Effect of Stepsize Selection}

We next examine how the choice of the stepsize exponent \(\beta\) affects the performance of our algorithm. Figure~\ref{fig:beta} shows the mean-squared error for different choices of \(\beta\in[0.01,0.99]\), with \(N=5\) agents and two levels of heterogeneity. Each curve is averaged over 10 independent runs of 5000 iterations. We observe that smaller values of \(\beta\) generally lead to faster error reduction in both heterogeneity settings. This is interesting because Theorem~\ref{thm : finite-time convergence} guarantees the same asymptotic convergence rate order for every \(\beta\in(0,1)\), while the experiments suggest that the finite-time performance can depend significantly on its choice. Determining an optimal choice of \(\beta\) remains an open question.

\subsection{Long-Term Behavior under Heterogeneity}

We finally examine the long-term behavior of the two algorithms in a heterogeneous environment. Figure~\ref{fig:heavyrun} shows their performance over 200,000 iterations with \(N=10\) agents and \(\epsilon_p=\epsilon_r=0.7\). We observe that the mean-squared error eventually levels off rather than converging to zero. This behavior is consistent with Theorem~\ref{thm : finite-time convergence}, which guarantees convergence to the federated solution \(\theta^*\), which may differ from the individual agent's solution \(\theta_1^*\). Thus, a nonzero error relative to \(\theta_1^*\) may persist even after the stochastic estimation error has diminished. Interestingly, our algorithm achieves a lower mean-squared error than the vanilla variant.

\section{Conclusions}
\label{sec:future}
We studied federated TD learning with linear function approximation in heterogeneous environments under the average-reward criterion. We proposed a two-timescale algorithm incorporating Polyak–Ruppert averaging. We showed that the algorithm achieves a mean-squared-error convergence rate of \(O(1/(NT))\), where \(N\) is the number of agents and \(T\) is the number of iterations, up to a residual error due to heterogeneity. Importantly, this rate is achieved using stepsizes that do not depend on unknown model parameters. Our results establish that a linear speedup in the number of agents is achievable even in heterogeneous environments without requiring problem-dependent stepsize tuning.


\bibliographystyle{ieeetr}
\bibliography{references}

\end{document}